\documentclass{article}

\usepackage[nonatbib, main, final]{neurips_2026}
\usepackage[utf8]{inputenc} % allow utf-8 input
\usepackage[T1]{fontenc}    % use 8-bit T1 fonts
\usepackage{hyperref}       % hyperlinks
\usepackage{url}            % simple URL typesetting
\usepackage{booktabs}       % professional-quality tables
\usepackage{amsfonts}       % blackboard math symbols
\usepackage{nicefrac}       % compact symbols for 1/2, etc.
\usepackage{microtype}      % microtypography
\usepackage{xcolor}         % colors

\usepackage[style=numeric, backend=biber]{biblatex} % thomai
\usepackage{graphicx}                               % thomai
\usepackage{multirow}                               % thomai
\usepackage{subcaption}                             % thomai

\usepackage{amsmath}
\usepackage{amssymb}
\usepackage{mathtools}
\usepackage{amsthm}
\theoremstyle{plain}
\newtheorem{theorem}{Theorem}[section]

\theoremstyle{definition}

\newtheorem{remark}[theorem]{Remark}

\usepackage{algorithm}
\usepackage{algorithmic}
\usepackage{wrapfig}

\title{Reformulating KV Cache Eviction Problem for Long-Context LLM Inference}

\author{%
  Tho Mai \\
  KAIST \\
  Daejeon, South Korea \\
  \texttt{thomh1511@kaist.ac.kr} \\
  \And
  Joo-Young Kim\thanks{Corresponding author.} \\
  KAIST \\
  Daejeon, South Korea \\
  \texttt{jooyoung1203@kaist.ac.kr} \\
}

\begin{document}

\maketitle

\begin{abstract}
  Large language models (LLMs) support long-context inference but suffer from substantial memory and runtime overhead due to Key-Value (KV) Cache growth. 
  Existing KV Cache eviction methods primarily rely on local attention weights, neglecting the influence of value representations, output projection, and inter-head interactions. 
  In this work, we reformulate KV Cache eviction from a conventional head-wise, weight-averaging approach into an output-aware, layer-wise matrix multiplication approximation problem. We introduce LaProx, a novel eviction strategy that explicitly models the multiplicative interaction between attention maps and projected value states to accurately quantify token contributions while accounting for inter-head dependencies.
  Building on this metric, we propose the first unified eviction strategy that assigns globally comparable importance scores to tokens, enabling model-wide selection instead of local, head-wise decisions.
  Experimental results across 19 datasets on long-context benchmarks LongBench and Needle-In-A-Haystack demonstrate that our approach maintains model performance with only 5\% of the KV cache and consistently outperforms prior works across all configurations. Notably, our method achieves up to 2$\times$ accuracy loss reduction under extreme compression scenarios compared to existing state-of-the-art baselines with minimal overhead. Our code is available at \url{https://github.com/huuthomai/LaProx}.
\end{abstract}

\section{Introduction}
Recent advances in large language models (LLMs) have significantly extended their ability to process long contexts, enabling strong performance in applications such as multi-turn dialogue \cite{vicuna2023}, question answering \cite{kamalloo-etal-2023-evaluating}, code generation \cite{Guo2023LongCoderAL}, and document understanding \cite{zhang-etal-2024-benchmarking}. 
To accelerate autoregressive inference, transformers cache key and value states from previous tokens, avoiding repeated attention computation. While this Key-Value (KV) cache is essential for efficient decoding, its size grows linearly with context length, quickly becomes a major bottleneck for memory usage and decoding latency in long-context settings. 
While techniques such as head merging or architectural modifications \cite{Ainslie2023GQATG} can partially alleviate these costs during training, they are often incompatible with fixed, pretrained models commonly used in deployment. Consequently, managing KV cache efficiently at inference time—without retraining or altering model parameters—becomes a critical challenge for scalable and cost-effective long-context LLM deployment under realistic memory and hardware constraints \cite{alisa}.

To operate large language models under constrained memory budgets, a common strategy is to dynamically reduce the size of the key-value (KV) cache by evicting entries deemed less influential during inference. 
Prior work has shown that, in practice, only a small subset of cached tokens meaningfully contributes to the attention output \cite{h2o, spatten}, motivating a class of eviction-based methods that selectively retain critical entries while discarding the rest. Early approaches exploit empirical observations that attention weights are highly concentrated, whereby a minority of tokens consistently receives the majority of attention mass. 
Building on this phenomenon, several methods identify important cache entries by averaging attention scores over time, with later refinements introducing observation windows, pooling mechanisms \cite{snapkv} or adaptive budget allocation \cite{cake} to better preserve salient information. 
However, such strategies are often heuristics and lack a principled formulation of what constitutes cache entry criticality. Consequently, the precise relationship between attention behavior, value representations, and their joint impact on the final model output remains insufficiently characterized.

In this paper, we reformulate KV cache eviction as an optimization problem that preserves layers’ attention outputs under a fixed budget. By explicitly modeling the output as a product of attention, value, and output matrix, we move beyond conventional attention-only heuristics, allowing us to rank cache entries by their actual contribution to the multiplicative interactions that form the final layer output. 
Critically, this formulation reveals that token importance is fundamentally coupled to the aggregate representation formed within each layer and, consequently, to the model’s eventual output. This observation suggests that eviction is most effectively managed at the model level rather than through isolated, head-wise decisions. 
Based on this insight, we propose a novel eviction strategy enabling more effective global cache selection.
% than existing head-wise attention-based methods.

Our contributions are summarized as follows:

1. We demonstrate that attention weights alone provide an incomplete measure of token importance, and accurate selection must account for output information and attention layer's structure itself.

2. We reveal that existing independent head-wise eviction is suboptimal because it neglects inter-head and inter-layer interactions, and show that eviction should be done at the model-level.

3. We introduce \textbf{La}yer \textbf{Approx}imated Cache (\textbf{LaProx}), a new eviction strategy that approximates layer's output by evaluating tokens across heads and layers simultaneously without any calibration.

4. Extensive evaluations on long-context benchmarks demonstrate that the proposed method consistently outperforms attention-based eviction strategies, confirming the effectiveness of our proposal.

\section{Background and Related Works}
\subsection{Basic of Attention and KV Cache Operations}
For clarity, we describe the mechanism using Multi-Head Attention (MHA) and omit the layer index, noting that the formulation applies identically to all transformer attention layers. 
Let $\mathbf{X} \in \mathbb{R}^{S \times D}$ be the token embeddings of a sequence of length $S$, where $D$ is the model hidden dimension. 
Each attention head operates on a subspace of dimension $d_h$, with $D = H \cdot d_h$ for $H$ heads. 
The projection matrices $\mathbf{W}_Q^{(h)}, \mathbf{W}_K^{(h)}, \mathbf{W}_V^{(h)} \in \mathbb{R}^{D \times d_h}$ map the shared hidden representations into head-specific query, key, and value states. During prompt processing, each head computes
\begin{equation}
    \mathbf{Q^{(h)}} = {\mathbf{X}\mathbf{W}_Q^{(h)}},  
    \mathbf{K^{(h)}} = {\mathbf{X}\mathbf{W}_K^{(h)}},  
    \mathbf{V^{(h)}} = {\mathbf{X}\mathbf{W}_V^{(h)}} \label{eq:qkv}
\end{equation}
with attention weights
\begin{equation}
    \mathbf{A^{(h)}} = \operatorname{Softmax}\left( \frac{Q^{(h)}{K^{(h)}}^\top}{\sqrt{d_h}} \right) \label{eq:attn}
\end{equation}
The per-head attention outputs are then concatenated,
\begin{equation}
    \mathbf{AV} = \operatorname{Concat}(\mathbf{A}^{(1)}\mathbf{V}^{(1)}, \dots, \mathbf{A}^{(h)}\mathbf{V}^{(h)})
    \label{eq:concat1}
\end{equation}
and projected to produce the final attention output,
\begin{equation}
    \mathbf{O} = \mathbf{AV}\mathbf{W}_O
    \label{eq:concat2}
\end{equation}
Following the projection $\mathbf{W}_O$, the final layer output is integrated via a residual connection:
\begin{equation}
\mathbf{Y} = \operatorname{Norm}(\mathbf{O} + \mathbf{X})\label{eq:residual}
\end{equation}
where $\mathbf{X}$ is the input identity and $\operatorname{Norm}$ denotes a normalization function.

During autoregressive decoding, at each decoding step $i$, only the newly generated token embedding $\mathbf{x_i} \in \mathbb{R}^{1 \times D}$ is projected to obtain its head-wise query, key, and value states. To avoid recomputation of past tokens, the new key-value pairs are appended to the cache
\begin{equation}
    \mathbf{K}^{(h)} \leftarrow \operatorname{Concat}(\mathbf{K}^{(h)}, \mathbf{x}_i \mathbf{W}_K^{(h)}), \qquad \mathbf{V}^{(h)} \leftarrow \operatorname{Concat}(\mathbf{V}^{(h)}, \mathbf{x}_i \mathbf{W}_V^{(h)})
\end{equation}
and the query $\mathbf{q_i^{(h)}} = {\mathbf{x_i}\mathbf{W}_Q^{(h)}}$ attends over the cached keys using equation \ref{eq:attn}.

While KV caching significantly reduces computation during decoding, the cache grows linearly with sequence length, leading to substantial memory and attention overhead in long-context inference.

\subsection{KV Cache Eviction}
KV cache eviction during inference reduces memory and computational overhead without modifying the attention mechanism. Its objective is to retain important tokens while removing low-impact ones. Early methods, such as StreamingLLM \cite{streamingllm}, adopt window-based strategies that preserve attention sinks and recent tokens while LongFormer \cite{longformer} uses two types of sliding windows cooperating with some pre-selected input locations. 
While efficient, these approaches may discard informative tokens in the middle of long sequences, degrading long-context performance. Other works, including H2O \cite{h2o} and Scissorhands \cite{scissorhands}, rank KV entries using accumulated attention scores to better capture token importance. 
Building on this line of work, SnapKV \cite{snapkv} and CAKE \cite{cake} further improve performance by averaging attention within an observation window and applying a pooling operation, achieving state-of-the-art (SOTA) results.

Beyond token selection, several studies explore non-uniform cache budget allocation. Layer-wise approaches such as PyramidInfer \cite{pyramidinfer} and PyramidKV \cite{pyramidkv} assign budgets based on network depth, while D2O \cite{d2o} and CAKE \cite{cake} adjust cache sizes using layer-specific attention variance. 
At the head level, AdaKV \cite{adakv} applies top-k selection across head-scores with an empirical safeguard, whereas HeadKV \cite{headkv} uses calibration procedures to determine fixed per-head budgets prior to inference.

A few works go beyond attention scores. For example, LAVa \cite{lava} and CAOTE \cite{caote} leverage value representations in their eviction indicators but omit the output projection; meanwhile, CriticalKV \cite{criticalkv} relies on two empirical safeguards to rescale the mean attention scores with output information, disregarding the actual formulation of the attention layer.

Despite their competitive results, existing methods rely primarily on attention weights for both eviction and budget allocation, or heuristically leverage output information \textbf{without considering the actual layer's formulation.} Furthermore, these approaches are limited to performing eviction on a per-head basis, \textbf{neglecting cross-head and cross-layer interactions.}
In contrast, this work proposes a principled eviction criterion that incorporates both attention probabilities and $VW_O$ contributions, and the cross-heads interaction, providing a more accurate measure of token importance.

\section{Motivation}
% \begin{wrapfigure}{r}{0.48\textwidth}
%   \centering
%   \vspace{-14pt} % Adjust to align top of figure with text
  
%   % --- Subfigure (a) ---
%   \begin{subfigure}{\linewidth}
%     \centering
%     \includegraphics[width=0.95\linewidth]{figures/Pattern.pdf}
%     \caption{$A$ and $VW_O$ patterns.}
%     \label{fig:pattern}
%   \end{subfigure}
  
%   % --- Subfigure (b) ---
%   \begin{subfigure}{\linewidth}
%     \centering
%     \includegraphics[width=0.95\linewidth]{figures/Magnitude.pdf}
%     \caption{Average strength.}
%     \label{fig:magnitude}
%   \end{subfigure}

%   \caption{Pattern and magnitude of $A$ and $VW_O$.}
%   \label{fig:total-comparison}
%   \vspace{-15pt} % Adjust to prevent large gaps below the figure
% \end{wrapfigure}

\begin{figure}[tb]
  \centering
  
  % --- Subfigure (a) ---
  \begin{subfigure}{0.48\textwidth}
    \centering
    \includegraphics[width=\linewidth]{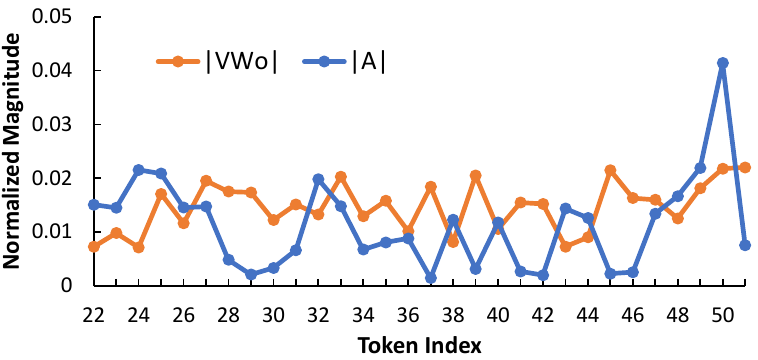}
    \caption{$A$ and $VW_O$ patterns.}
    \label{fig:pattern}
  \end{subfigure}
  \hfill
  % --- Subfigure (b) ---
  \begin{subfigure}{0.48\textwidth}
    \centering
    \includegraphics[width=\linewidth]{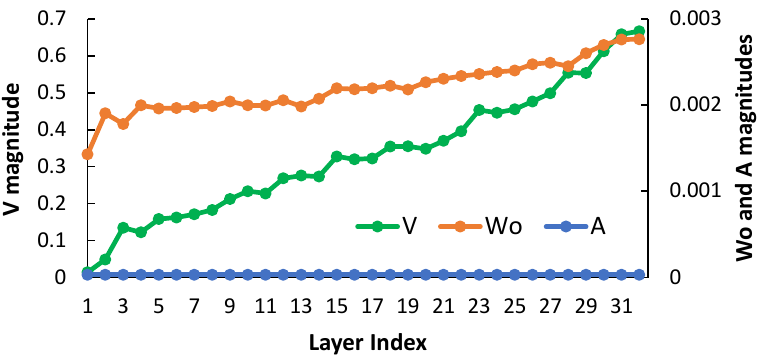}
    \caption{Average strength.}
    \label{fig:magnitude}
  \end{subfigure}

  \caption{Pattern and magnitude of $A$ and $VW_O$.}
  \label{fig:total-comparison}
\end{figure}

In this section, we investigate the relationship between attention weight ($A$) and the value–output projection ($VW_O$). Specifically, we examine whether the average of $A$ alone can serve as a faithful proxy for the attention layer output, i.e., whether attention weights $A$ are sufficient to characterize the whole product $AVW_O$. This approach assumes two key conditions are satisfied: (1) the patterns of $A$ and $VW_O$ are well aligned, and (2) the magnitude of $A$ is not dominated by $VW_O$.

\textbf{Experiment setup.} Our analysis is conducted using the Mistral-7B-Instruct-v0.3 model. For visualization clarity, we display only a contiguous subset of tokens.

\textbf{Observation.} Figure \ref{fig:pattern} reports the normalized per-token magnitudes of $|A|$ and $|VW_O|$. While the two quantities share some high-score tokens (such as \#25 or \#49-50), their overall patterns differ significantly. Many steps even show opposite peaks; for instance, tokens \#37 and \#39 have high $VW_O$ values but low $A$ values. This indicates that $A$ and $VW_O$ assess token importance differently, and one cannot be used in place of the other.

Furthermore, Figure \ref{fig:magnitude} reveals that the value range of $A$ is much smaller than $VW_O$. As attention weights are normalized probabilities, their values are confined to a narrow range, whereas $VW_O$ has a much wider range of values, which expands in deeper layers.

These observations demonstrate that attention weights alone are insufficient to represent the attention layer output, motivating the incorporation of value and output projection in cache eviction decisions.

\section{Methodology}
\subsection{Eviction Indicator}
\label{sec:eviction-score}
Equations \ref{eq:concat1} and \ref{eq:concat2} show that the standard MHA is defined as the concatenation of all head outputs followed by a linear projection. Although the output projection mixes attention information from all heads, the computation can be exactly decomposed into a sum of independent head-wise contributions. 
\begin{remark}
\label{rem:decompose}
Let \(W_O = [W_O^1, W_O^2, \dots, W_O^h] \in \mathbb{R}^{h d_h \times d}\)
be the output projection matrix partitioned along the head dimension, and \(H^i=A^iV^i \in \mathbb{R}^{S \times d_h}\) denote the output of the \(i\)-th attention head.
Then, the output of a standard MHA layer can be expressed as
\begin{equation}
\text{Output} = \sum_{i=1}^{h} H^i W_O^i = \sum_{i=1}^{h} A^i V^i W_O^i.
\label{eq:attn_decomp}
\end{equation}
\end{remark}
% \textit{Proof.}
% See Appendix \ref{proof:MHA-decomposition} for details. \hfill$\square$

\begin{wrapfigure}{r}{0.52\textwidth} % {r} for right side, 0.5 for half-width
  \vspace{-20pt} % Adjust vertical space to align with text
  \begin{minipage}{\linewidth}
% \begin{algorithm}[tb]
\begin{algorithm}[H]
\caption{Eviction Score Computation}
\label{alg:kv_eviction}
\begin{algorithmic}
\STATE {\bfseries Input:} Query $\mathbf{Q}$, KV Cache $(\mathbf{K}, \mathbf{V})$, Projection $W_O$, Budget $B_{total}$, Observation Window $w$
\STATE {\bfseries Output:} Compressed KV cache $(\tilde{\mathbf{K}}, \tilde{\mathbf{V}})$

\vspace{0.5em}
\STATE {\textcolor{green!50!black}{// Compute attention weight and projected values}}
\STATE $\mathbf{A} \leftarrow \operatorname{Softmax}\!\left(\frac{\mathbf{Q[-w:,]}\mathbf{K}^\top}{\sqrt{d_k}}\right)$
\STATE $\mathbf{H} \leftarrow V W_O$

\vspace{0.5em}
\STATE {\textcolor{green!50!black}{// Score tokens}}
\STATE $T \leftarrow$ number of cached tokens
\FOR{$i=0$ {\bfseries to} $T$}
\IF{$i < T-w$}
\STATE $\mathbf{p[i]} \leftarrow \left\|\mathbf{A[:,i]}\right\|_{2} \cdot \left\|\mathbf{H[i,:]}\right\|_{2}$
\ELSE
\STATE $\mathbf{p[i]} \leftarrow \infty$
\ENDIF
\ENDFOR
\vspace{0.5em}
\STATE {\textcolor{green!50!black}{// Evict tokens}}
% \IF{$B_{total} \ge T$}
% \STATE $(\tilde{\mathbf{K}}, \tilde{\mathbf{V}}) \leftarrow (\mathbf{K}, \mathbf{V})$
% \ELSE
\STATE $\mathcal{S} \leftarrow \operatorname{TopK}(\mathbf{p}, B_{total})$
\STATE $(\tilde{\mathbf{K}}, \tilde{\mathbf{V}}) \leftarrow (\mathbf{K}[\mathcal{S}], \mathbf{V}[\mathcal{S}])$
% \ENDIF
\vspace{0.5em}
\STATE \textbf{return} $(\tilde{\mathbf{K}}, \tilde{\mathbf{V}})$
\end{algorithmic}
\end{algorithm}
\end{minipage}
  \vspace{-10pt}
\end{wrapfigure}

Remark \ref{rem:decompose} shows that \textbf{by integrating $\boldsymbol{VW_O}$ at the head level, we can evaluate token importance across the entire layer}. To quantify token importance, we leverage matrix multiplication associativity and compute $VW_O$ first to preserve the key-value alignment between the attention matrix $A$ and the projected values $VW_O$.
A naive but limited approach of independently \textbf{scaling the average attention scores by the magnitude of $VW_O$ will overlook the fundamental nature of the attention output,} which is formed through a matrix multiplication. 
Since our objective is to preserve this layer attention output under a constrained KV cache budget, we instead treat cache eviction as a matrix multiplication approximation problem: selecting a subset of tokens that best approximates the full product $A \times (VW_O)$.
From this perspective, we draw on Monte Carlo analyses of matrix multiplication to provide a rigorous mathematical basis for our eviction criteria. This theory states that the approximation error of a matrix product is minimized when the selection follows the product of the Euclidean norms of the corresponding column-row pairs \cite{montecarlo,crs}. 
\begin{equation}
    % p_i = \frac{\|\boldsymbol{A}_{:,i}\|_2 \|\boldsymbol{VWo}_{i,:}\|_2}{\sum_{j}^S \|\boldsymbol{A}_{:,j}\|_2 \|\boldsymbol{VWo}_{j,:}\|_2}
    p_i = \frac{\|\boldsymbol{A}[:,i]\|_2 \ \|\boldsymbol{VW_O}[i,:]\|_2}{\sum_{j}^S \|\boldsymbol{A}[:,j]\|_2 \ \|\boldsymbol{VW_O}[j,:]\|_2}
    \label{eq:evict_prob}
\end{equation}
More generally, the score can be represented by:
\begin{equation}
    % p_i \propto \|\boldsymbol{A}_{:,i}\|_2 \|\boldsymbol{VWo}_{i,:}\|_2
    p_i \propto \|\boldsymbol{A}[:,i]\|_2 \ \|\boldsymbol{VWo}[i,:]\|_2
    \label{eq:evict_score}
\end{equation}
 Equation \ref{eq:evict_score} shows the eviction score $\mathbf{p_i}$ per token $i$ per head.
 % , where $\mathbf{A}$ is the attention probability matrix, $\mathbf{VW_o}$ is our Value-Output matrix, and $\mathbf{S}$ is the sequence length. 
 Intuitively, this criterion favors indices that simultaneously carry significant mass in both matrices, which are the terms that dominate the output. In this work, we select top indices with the highest scores given by equation \ref{eq:evict_score} to ensure that the most influential tokens are preserved. 
 By grounding our eviction indicator in this matrix approximation principle, we obtain a token importance measure that directly \textbf{aligns with the structure of the attention computation} and more faithfully preserves the layer output compared to heuristics based solely on attention weights. The detailed algorithm is shown in Algorithm \ref{alg:kv_eviction}.

\subsection{Eviction Action}
The additive structure in equation \ref{eq:attn_decomp} implies that, although we estimate token impact independently within each head, the resulting contribution of a token is not confined to that head alone; rather, it directly participates in the formation of the shared layer output after the output projection. Consequently, token importance is fundamentally a layer-level concept: tokens with large estimated contributions impact the layer output equivalently, regardless of which head produces them.

In contrast, head-wise selection relies on head local rankings, which can be misleading. For example, a token may rank highly in its head, but its projected contribution after $VW_O$ can be \textbf{numerically insignificant compared to tokens from other heads and provides minimal difference in the aggregation step.} Retaining such “local winners” may waste memory on signals that have little impact on the final layer output. Conversely, some tokens may not be top-ranked in their own heads, yet contribute highly across multiple heads. These tokens are naturally captured by a layer-level criterion but missed by head-wise selection.

\begin{remark}
\label{rem:layer_token}
For a multi-head attention layer $l$, the contribution of token $j$ at query position $i$ is
\begin{equation}
\Delta \mathbf{o}^l(i,j)
=
\sum_{h'=1}^{h}
A^{l,h'}(i,j)\,
V^{l,h'}(j)\,
W_{O}^{l,h'}
\label{eq:token_contribution1}
\end{equation}
and the layer output satisfies
\begin{equation}
\mathbf{o}^l(i)
=
\sum_{j}
\Delta \mathbf{o}^l(i,j).
\label{eq:layer_output}
\end{equation}
Therefore, the contribution of a token from any head enters the layer output directly through linear addition. Token importance must therefore be evaluated at the layer level.
\end{remark}

% \textit{Proof.}
% See Appendix \ref{appendix:layer-importance} for details. \hfill$\square$

\textbf{With the presence of $\boldsymbol{VW_O}$}, remark \ref{rem:layer_token} motivates formulating KV cache eviction as a unified selection problem across an entire layer, rather than a series of independent head-wise decisions. Unlike existing weight-based methods \cite{adakv} which rely on eviction scores that are strictly local to each head, our metric (Section \ref{sec:eviction-score}) integrates $W_O$ that maps head contributions onto a unified scale, making token importance comparable across the entire layer, and providing a theoretically grounded basis for global selection. 
This allows us to safely flatten eviction scores and perform a joint selection under a fixed layer budget, and naturally enables an adaptive allocation of cache capacity: heads containing influential tokens retain more entries, while those with lower-impact tokens are pruned more aggressively. In Table \ref{tab:score-alloc}, we also prove that a greedy selection without our global score can even hurt the performance.

Similar to how a layer output is formed by accumulating contributions from tokens across all heads, \textbf{the final model output is also an accumulation across layers.} As shown in Equation \ref{eq:residual}, a Transformer consists of stacked blocks connected through a shared residual path, where each layer’s output is added to this path. As a result, the final output reflects the aggregated contributions of all layers. Therefore, if a token has a strong impact on its layer’s output, this impact is directly propagated through the residual path and influences the final output. In this sense, a token’s importance can be understood by how much it affects/changes this accumulated representation.

\begin{wrapfigure}{r}{0.34\textwidth} % {r} for right side, 0.5 for half-width
  \vspace{-10pt} % Adjust vertical space to align with text
  \begin{minipage}{\linewidth}
% \begin{algorithm}[tb]
\begin{algorithm}[H]
  \caption{Eviction Action}
  \label{alg:budget_allocation}
  \begin{algorithmic}
    \STATE {\bfseries Input:} Eviction scores $\{p_{l,h,j}\}$, global budget $K$
    \STATE {\bfseries Output:} Selected token set $\mathcal{S}$

    \vspace{0.5em}
    \STATE {\textcolor{green!50!black}{// Flatten head-wise scores}}
    \FOR{each layer $l$}
      \FOR{each head $h$}
        \FOR{each token $j$}
          \STATE $p_{l,k} \leftarrow p_{l,h,j}$
        \ENDFOR
      \ENDFOR
    \ENDFOR

    \vspace{0.5em}
    \STATE {\textcolor{green!50!black}{// Layer-wise normalization}}
    \FOR{each layer $l$}
      \FOR{each token $j$}
        \STATE $s_{l,j} \leftarrow p_{l,j} / \sum_k p_{l,k}$
      \ENDFOR
    \ENDFOR

    \vspace{0.5em}
    \STATE {\textcolor{green!50!black}{// Global selection}}
    \STATE $\mathcal{S} \leftarrow \operatorname{TopK}\!\left(\{s_{l,j}\}_{l,j}, K\right)$
  \end{algorithmic}
\end{algorithm}
\end{minipage}
  \vspace{-10pt}
\end{wrapfigure}

However, while our $VW_O$-based metric accurately captures a token's contribution to the layer output, these raw scores are not directly comparable across different layers due to the inherent magnitude variations shown in Figure \ref{fig:magnitude}. To resolve this scale disparity, we employ a simple layer-wise score normalization scheme that transforms flattened scores into a relative importance distribution:
\begin{equation}
s_{l,j} = \frac{p_{l,j}}{\sum_{k} p_{l,k}},
\label{eq:layer_norm}
\end{equation}
where $p_{l,j}$ denotes the flattened eviction score of token $j$ in layer $l$. This normalization neutralizes inter-layer scale differences for inter-layer comparison while strictly preserving the relative token rankings within each layer. The effectiveness of this simple normalization and its necessity is further demonstrated empirically in our evaluation and ablation study in Section \ref{sec:experiments} and Appendix \ref{sec:ablation}.

Finally, tokens across the model are jointly selected using {${s_{l,j}}$} (Algorithm~\ref{alg:budget_allocation}). \textbf{This unified rule allows the model to focus on the most critical tokens model-wide.} Notably, this approach is \textit{method-specific hyperparameter-free} and requires no calibration or training/finetuning, offering a straightforward implementation without complexity.

\section{Experiments}
\label{sec:experiments}

\textbf{Models.} We conduct experiments using three widely-used open-sourced LLMs: Llama-3.1-8B-Instruct \cite{llama3}, Mistral-7B-Instruct-v0.3 \cite{mistral7b}, and Qwen3-8B \cite{qwen3}. 
These models provide maximum context lengths of 128K, 32K, and 32K tokens, respectively. Additionally, we also perform experiments on larger models in Appendix \ref{appendix:moe} to illustrate our approach's scalability.

\textbf{Baselines.} Our method is compared against the full-cache configuration (FullKV) and four SOTA baselines: StreamingLLM (SLLM) \cite{streamingllm}, SnapKV \cite{snapkv}, AdaKV \cite{adakv}, 
CriticalKV \cite{criticalkv}, 
and CAKE \cite{cake}. 
A summary of these baselines is provided in Table \ref{table-baseline}.

\textbf{Evaluation Scenarios.} Our approach is compared across various memory budgets (average capacity per head). 
We use fixed absolute cache sizes rather than ratios relative to the full KV cache to prevent cache size increase with context length, better reflecting real-world hardware constraints. 
For implementation details, see Appendix \ref{appendix:imp}.

\textbf{Evaluation Benchmarks.} Following the standard practice of the prior works, our main experiments include two long-context benchmarks: LongBench \cite{longbench} and RULER \cite{ruler}. Additionally, we also evaluate on InfiniteBench \cite{infinitebench}, a very-long-context benchmark in \ref{sec:infinitebench}.

\subsection{Evaluations on LongBench Dataset}
\begin{figure*}[!t]
  \begin{center}
    % Change \columnwidth to \textwidth
    % \centerline{\includegraphics[width=\textwidth]{figures/LongBench.pdf}}
    \centerline{\includegraphics[width=\textwidth]{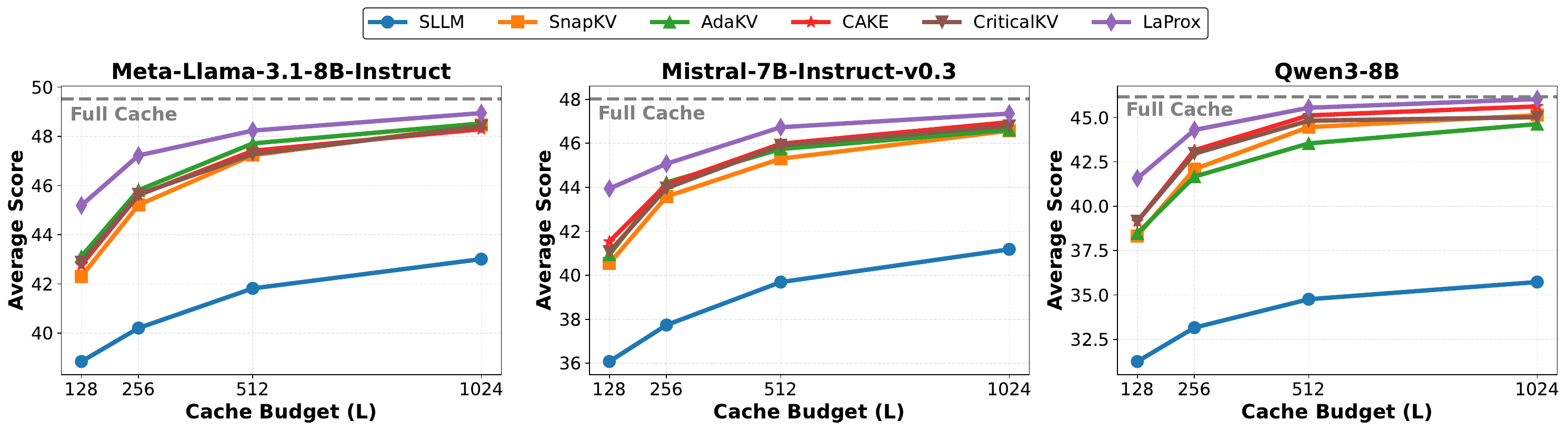}}
    \caption{
      Average scores among 16 datasets of LongBench under different cache budgets.
    }
  \label{fig:longbench}
  \end{center}
\end{figure*}

\begin{table*}[!t]
% \caption{Comparison across 16 LongBench datasets. The best result is highlighted in \textbf{bold} and the second best in \underline{underline}.}
\caption{Comparison across 16 LongBench datasets, with \textbf{best} and \underline{second best} results highlighted.}
\label{table-longbench}
\centering
\small
\setlength{\tabcolsep}{3pt}
\resizebox{\textwidth}{!}{
\begin{tabular}{l ccc ccc ccc ccc cc cc c}
\toprule
\multirow{2}{*}{Method} & \multicolumn{3}{c}{Single-Document QA} & \multicolumn{3}{c}{Multi-Document QA} & \multicolumn{3}{c}{Summarization} & \multicolumn{3}{c}{Few-shot Learning} & \multicolumn{2}{c}{Synthetic} & \multicolumn{2}{c}{Code} & \multirow{2}{*}{Avg.} \\
\cmidrule(lr){2-4} \cmidrule(lr){5-7} \cmidrule(lr){8-10} \cmidrule(lr){11-13} \cmidrule(lr){14-15} \cmidrule(lr){16-17}

 & NrtvQA & Qasper & MF-en & HotpotQA & 2WikiMQA & Musique & GovRep & QMSum & MultiNews & TREC & TriviaQA & SAMSum & PCount & PR-en & Lcc & RB-P &  \\
\midrule
\multicolumn{18}{c}{Meta-Llama-3.1-8B-Instruct, $B_{\text{total}} = 128L$} \\
\midrule
FullKV & 30.28 & 45.46 & 54.94 & 56.02 & 46.5 & 31.28 & 35.11 & 25.28 & 27.19 & 72.5 & 91.65 & 43.86 & 8.88 & 99.5 & 65.13 & 58.7 & 49.51 \\
SLLM & 22.45 & 20.83 & 31.74 & 44.86 & 39.31 & 23.88 & 18.75 & 20.78 & 18.57 & 40.5 & 85.57 & 38.24 & 8.16 & \textbf{99.5} & 58.89 & 49.7 & 38.86 \\
SnapKV & 25.53 & 23.92 & 48.15 & \underline{52.94} & 43.5 & 26.67 & 19.48 & 22.47 & 19.68 & 47 & 89.56 & 39.51 & \underline{8.41} & 98 & 59.43 & 52.84 & 42.31 \\
AdaKV & 25.74 & 24.8 & 49.38 & 52.73 & 43.75 & 27.8 & 20.37 & \underline{23.02} & \underline{20.38} & \underline{49} & \textbf{90.19} & 40.12 & 8.38 & \underline{99} & \underline{60.92} & \underline{54.4} & \underline{43.12} \\
CAKE & \textbf{25.93} & 25.5 & 48.33 & 51.95 & 43.42 & \underline{28.94} & \underline{20.91} & 22.93 & 19.96 & 47.5 & 88.84 & \textbf{41.26} & 8.38 & \underline{99} & 59.56 & 51.97 & 42.77 \\
CriticalKV & 24.56 & \underline{25.89} & \underline{49.86} & 52.42 & \underline{43.89} & 27.96 & 20.71 & 22.69 & 20.06 & 48.5 & 89.61 & 39.94 & 8.38 & \underline{99} & 60.08 & 52.56 & 42.88 \\
LaProx & \underline{25.78} & \textbf{31.38} & \textbf{52.97} & \textbf{55.17} & \textbf{44.95} & \textbf{29.46} & \textbf{21.66} & \textbf{23.66} & \textbf{21.47} & \textbf{59} & \underline{89.94} & \underline{41.09} & \textbf{8.89} & \textbf{99.5} & \textbf{61.91} & \textbf{56.16} & \textbf{45.19} \\
\midrule
\multicolumn{18}{c}{Mistral-7B-Instruct-v0.3, $B_{\text{total}} = 128L$} \\
\midrule
FullKV       & 29.07 & 41.58 & 52.88 & 49.37 & 39.01 & 28.58 & 34.81 & 25.66 & 27.82 & 76 & 88.59 & 47.4 & 5.5 & 98 & 61.4 & 62.53 & 48.01 \\
SLLM & 21.42 & 22.28 & 26.82 & 37.28 & 33.31 & 17.61 & 16.73 & 19.77 & 17.86 & 45.5 & 85.64 & 40.47 & 5.5 & 80 & 55.13 & 52.07 & 36.08 \\
SnapKV       & 23.81 & 25.05 & 46.92 & 44.98 & 35.06 & 23.11 & 20.49 & 21.55 & 19.32 & 43.5 & \textbf{89.78} & 42.87 & 6 & 93.5 & 55.95 & 54.6 & 40.41 \\
AdaKV        & 24.08 & 25.52 & 47.08 & 47.16 & 35.16 & 23.83 & 20.23 & 22.28 & 20.7 & 45 & 88.92 & \textbf{43.77} & \underline{6.5} & 92.5 & \underline{56.63} & \underline{55.81} & 40.95 \\
CAKE         & 25.04 & \underline{27.32} & \underline{48.49} & \underline{47.48} & 35.24 & \underline{24.63} & \textbf{21.32} & \underline{22.54} & \underline{20.78} & 45.5 & 89.41 & 43.37 & \textbf{7} & \underline{95} & 56.47 & 54.99 & \underline{41.54} \\
CriticalKV   & \underline{25.42} & 25.95 & 46.43 & 46.44 & \underline{35.92} & 23.93 & 21.12 & 22.35 & 19.26 & \underline{47} & 89.23 & 43.62 & 6.5 & 93.5 & 55.99 & 54.51 & 41.07 \\
LaProx         & \textbf{28} & \textbf{30.26} & \textbf{54.02} & \textbf{47.86} & \textbf{37.54} & \textbf{25.54} & \underline{21.26} & \textbf{23.31} & \textbf{22.1} & \textbf{63.5} & \underline{89.7} & \underline{43.46} & 5 & \textbf{96} & \textbf{58.85} & \textbf{57.66} & \textbf{44.00} \\

\midrule
\multicolumn{18}{c}{Qwen3-8B, $B_{\text{total}} = 128L$} \\
\midrule
FullKV       & 32.26 & 46.59 & 52.73 & 59.35 & 51.21 & 33.2 & 32.44 & 24.13 & 25.68 & 71 & 65.82 & 40 & 1 & 100 & 53.06 & 50.17 & 46.17 \\
SLLM          & 12.18 & 28.05 & 21.79 & 36.34 & 39.07 & 6.88 & 15.78 & 18.54 & 15.95 & 43 & 62.56 & 33.3 & \underline{3} & 73 & 45.16 & 45.43 & 31.25 \\
SnapKV       & 16.34 & 31.35 & 42.37 & 52.67 & 42.33 & 18.26 & 16.03 & 19.68 & 16.49 & 53 & \underline{75.69} & 34.44 & 1 & \textbf{99} & 46.86 & 47.86 & 38.33 \\
AdaKV        & \textbf{21.66} & 32.58 & 43.96 & 50.81 & 42.95 & \underline{20.9} & 16.45 & 19.33 & 16.02 & 51 & 73.92 & 34.24 & 1 & \underline{98} & 45.77 & 46.92 & 38.47 \\
CAKE         & \underline{19.69} & 32.67 & 45.07 & \underline{57.5} & \textbf{48.02} & 20.5 & \underline{17.82} & \underline{20.87} & 16.73 & 47 & 69.51 & \underline{35.39} & \textbf{5} & \textbf{99} & 45.55 & 45.94 & 39.14 \\
CriticalKV   & 18.47 & \underline{35.83} & \underline{45.78} & 48.59 & 42.84 & 19.09 & 16.67 & 19.94 & \textbf{17.61} & \underline{56} & 75.31 & 35.3 & 0 & \textbf{99} & \underline{47.79} & \underline{48.28} & \underline{39.15} \\
LaProx         & 17.51 & \textbf{36.91} & \textbf{51.22} & \textbf{58.64} & \underline{46.31} & \textbf{23.85} & \textbf{18.68} & \textbf{21.21} & \underline{17.32} & \textbf{61} & \textbf{76.49} & \textbf{35.86} & 1 & \textbf{99} & \textbf{48.86} & \textbf{51.21} & \textbf{41.57} \\
\bottomrule
\end{tabular}
}
% \vskip -0.1in
\end{table*}

We evaluate LaProx against SOTA KV cache eviction techniques across the 16 datasets in the LongBench benchmark, using cache budgets ranging from 128 to 1024 tokens. Table \ref{table-longbench} details the performance across three models at a budget of 128 tokens, while Figure \ref{fig:longbench} illustrates the average performance across the full range of budget constraints.

As shown in Table \ref{table-longbench}, LaProx consistently outperforms previous works in nearly every LongBench's dataset, leading to significant improvements in total performance. Figure \ref{fig:longbench} further demonstrates our superior results across all budget sizes and models. 
Notably, the performance gap between LaProx and the baselines widens as the memory budget becomes more constrained, highlighting the robustness of LaProx under extreme hardware limitations.

We can also see that SLLM consistently exhibits the weakest performance, which is expected given its aggressive removal of intermediate tokens. 
By contrast, other approaches improve the performance by actively selecting important tokens. 
However, these baselines introduce only heuristic refinements to SnapKV, lacking a principled foundation or consideration of the underlying model structure, thereby limiting their gains over the vanilla SnapKV. 
Nevertheless, they remain strong baselines due to their improvements across many settings.
In some cases, however, \textbf{such heuristic strategies can even degrade performance,} as observed for Qwen3 at budgets of 256-1024 by AdaKV and CriticalKV.
Meanwhile, LaProx achieves superior performance across the majority of datasets and memory configurations. These results underscore the advantages of our unified global eviction strategy.

\subsection{Evaluations on Needle-in-A-Haystack}

\begin{figure*}[tb]
  \begin{center}
    % Change \columnwidth to \textwidth
    % \centerline{\includegraphics[width=\textwidth]{figures/NIAH.pdf}}
    \centerline{\includegraphics[width=\textwidth]{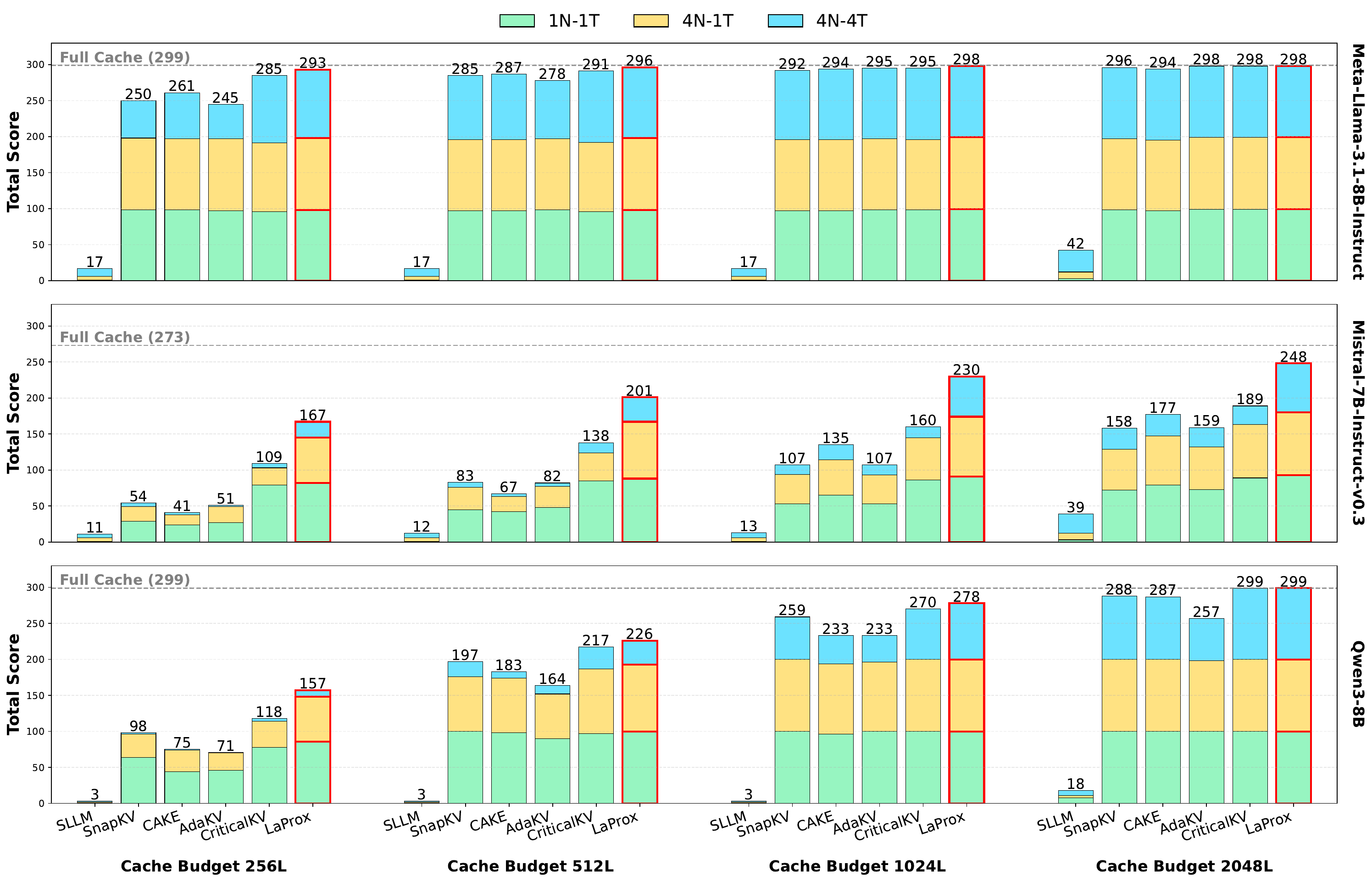}}
    \caption{
      Comparison across 3 NIAH variants at 32K context length.
    }
  \label{fig:niah}
  \end{center}
\end{figure*}

To evaluate retrieval performance, we employ Needle-in-A-Haystack (NIAH) test\nocite{needle}, where a target sentence is embedded within a long-context distractor. 
Following RULER \cite{ruler}, we examine three representative configurations: (1) Single-Needle with 1 needle and 1 target (1N-1T): There is a single needle that the model must retrieve from the context; (2) Multi-Needle with 4 needles and 1 target (4N-1T): 4 needles (1 target and 3 distractors) are inserted to the context and the model must isolate a specific target from three distracting needles; and (3) Multi-Needle with 4 needles and 4 targets (4N-4T): 4 needles are inserted and the model must retrieve all of them. 
To match the Mistral and Qwen models’ context windows and to balance the evaluation cost, each configuration is evaluated at a 32K context length across 100 samples per task, using cache budgets ranging from 256 to 2048 tokens. Further evaluations on RULER tasks are provided in Appendix \ref{appendix:agnostic} and \ref{appendix:moe}.

Consistent with LongBench results, LaProx demonstrates significant performance gains on NIAH tests across all evaluated models and cache budgets. Notably, on Mistral-7B-Instruct-v0.3 with a highly constrained budget of 256 tokens, LaProx achieves a 1.5 to 3$\times$ improvement in retrieval accuracy over CriticalKV and other prior methods, respectively, across all three NIAH variants.

Meanwhile, although the baselines perform more competitively on Llama-3.1-8B-Instruct and Qwen3-8B, LaProx consistently maintains the highest scores and reaches FullKV performance sooner than other approaches. The performance gap becomes particularly pronounced in resource-constrained settings (256 tokens) or complex tasks (4 needles).

Furthermore, similar to the LongBench observations, \textbf{heuristic refinement methods continue to exhibit unstable behavior} and, in many cases, even degrade performance relative to the vanilla SnapKV baseline (such as CAKE and AdaKV).

\subsection{Analysis of Matrix-Approximation-based Eviction Criterion}
\begin{wrapfigure}{r}{0.5\textwidth} % {r} for right side, 0.5 for half-width
  \vspace{-22pt} % Adjust vertical space to align with text
  \begin{minipage}{\linewidth}
\begin{figure}[H]
  % \vskip 0.2in
  \begin{center}
    \centerline{\includegraphics[width=\columnwidth]{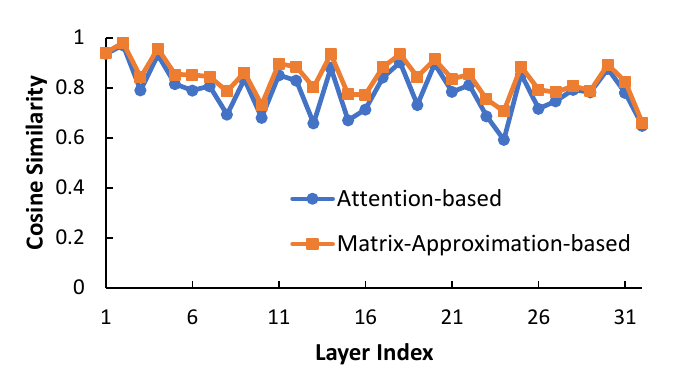}}
    \caption{
      Similarity score between the full and approximated attention layer outputs
    }
  \label{fig:cosine}
  \end{center}
\end{figure}
\end{minipage}
  % \vspace{-10pt}
\end{wrapfigure}

Beyond benchmark accuracy, we further investigate whether our matrix-based eviction criterion improves the similarity between the full and approximated attention outputs. Specifically, for each layer, we measure the cosine similarity between the full and compressed attention outputs for the first decoding token in Mistral-7B-Instruct-v0.3. 
The evaluation is conducted on the TREC dataset, with the KV cache of 128 tokens. As shown in Figure~\ref{fig:cosine}, our method consistently achieves higher cosine similarity across all layers compared to the vanilla approach based solely on local attention weights, indicating a more faithful approximation of the attention output.

\subsection{Evaluations on Efficiency}
\label{sec:efficiency}
% \begin{wrapfigure}{r}{0.45\textwidth}
%   \centering
%   \vspace{-15pt} % Adjust top spacing to align with text
  
%   % --- Plot (a) ---
%   % Scaled to fit within the wrapfigure's 0.5\textwidth
%   \begin{subfigure}{\linewidth}
%     \centering
%     \includegraphics[width=0.9\linewidth]{figures/Latency_Prefill.pdf}
%     \caption{Prefilling latency.}
%     \label{fig:prefill}
%   \end{subfigure}

%   % --- Plot (b) ---
%   \begin{subfigure}{\linewidth}
%     \centering
%     \includegraphics[width=0.9\linewidth]{figures/Latency_Decode.pdf}
%     \caption{Decoding latency.}
%     \label{fig:decode}
%   \end{subfigure}

%   \caption{Latency analysis.}
%   \label{fig:total-comparison}
%   \vspace{-10pt} % Adjust bottom spacing
% \end{wrapfigure}

\begin{figure*}[!t]
  \begin{center}
    % Change \columnwidth to \textwidth
    % \centerline{\includegraphics[width=\textwidth]{figures/LongBench.pdf}}
    \centerline{\includegraphics[width=\textwidth]{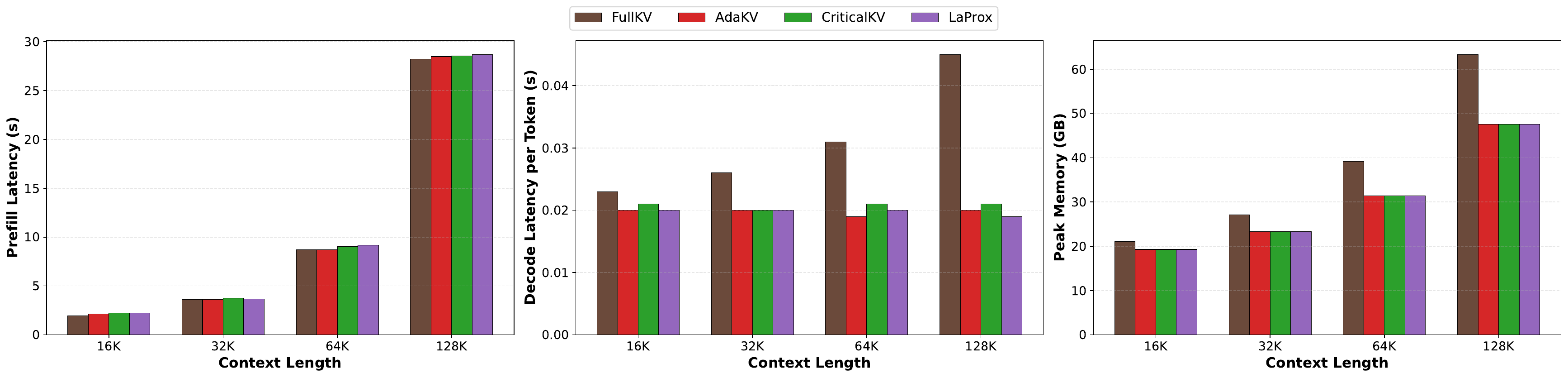}}
    \caption{
      Efficiency Analysis.
    }
  \label{fig:efficiency}
  \end{center}
\end{figure*}

To evaluate efficiency, we report both prefill latency, which includes initial prompt processing and eviction overhead, per-token decoding latency, and peak memory usage. All measurements are conducted on a single NVIDIA H100 (80GB) GPU using the Meta-Llama-3.1-8B-Instruct model with a fixed cache budget of 128 tokens. In this section, we compare our approach with AdaKV and CriticalKV, the strongest baselines supporting budget-allocation-based and output-aware eviction, respectively. Further efficiency analyses, including multibatch settings and latency breakdowns, can also be found in Appendix \ref{appendix:efficiency}

As shown in Figure \ref{fig:efficiency}, our method introduces only marginal overhead during prefilling across all context lengths. During decoding, all eviction methods achieve comparable efficiency and consistently outperform the FullKV setting. In contrast to FullKV, whose decoding latency grows rapidly with sequence length, our approach maintains a stable per-token latency by enforcing a strict KV cache budget. Consequently, at 128K context length, our method achieves a 2.3$\times$ decoding speedup over FullKV.

In addition, all eviction methods substantially reduce peak memory usage compared to FullKV. For instance, at 128K context length, eviction-based approaches require only around 47.5GB of memory, whereas FullKV consumes 63.3GB.

\subsection{Analysis on the Number of Retained Tokens}
Implicitly, our model-wide eviction strategy (Algorithm \ref{alg:budget_allocation}) retains different numbers of tokens across layers and heads according to their estimated importance. To analyze this behavior, we conduct experiments on the NIAH dataset using two representative models, Meta-Llama-3.1-8B-Instruct (Llama) and Mistral-7B-Instruct-v0.3 (Mistral), and record the number of retained tokens for each head. The total cache size is fixed to 128 tokens, corresponding to an average of 96 retained entries per head (with an additional 32-entry history window).

% \begin{figure*}[t]
%   \begin{center}
%     \centerline{\includegraphics[width=\textwidth]{figures/Budget_sq.pdf}}
%     \caption{
%       Average Budget Allocation and Variation per Head
%     }
%   \label{fig:budget}
%   \end{center}
% \end{figure*}
\begin{wrapfigure}{r}{0.5\textwidth} % {r} for right side, 0.5\textwidth for width
  \centering
  \vspace{-10pt}
  \includegraphics[width=\linewidth]{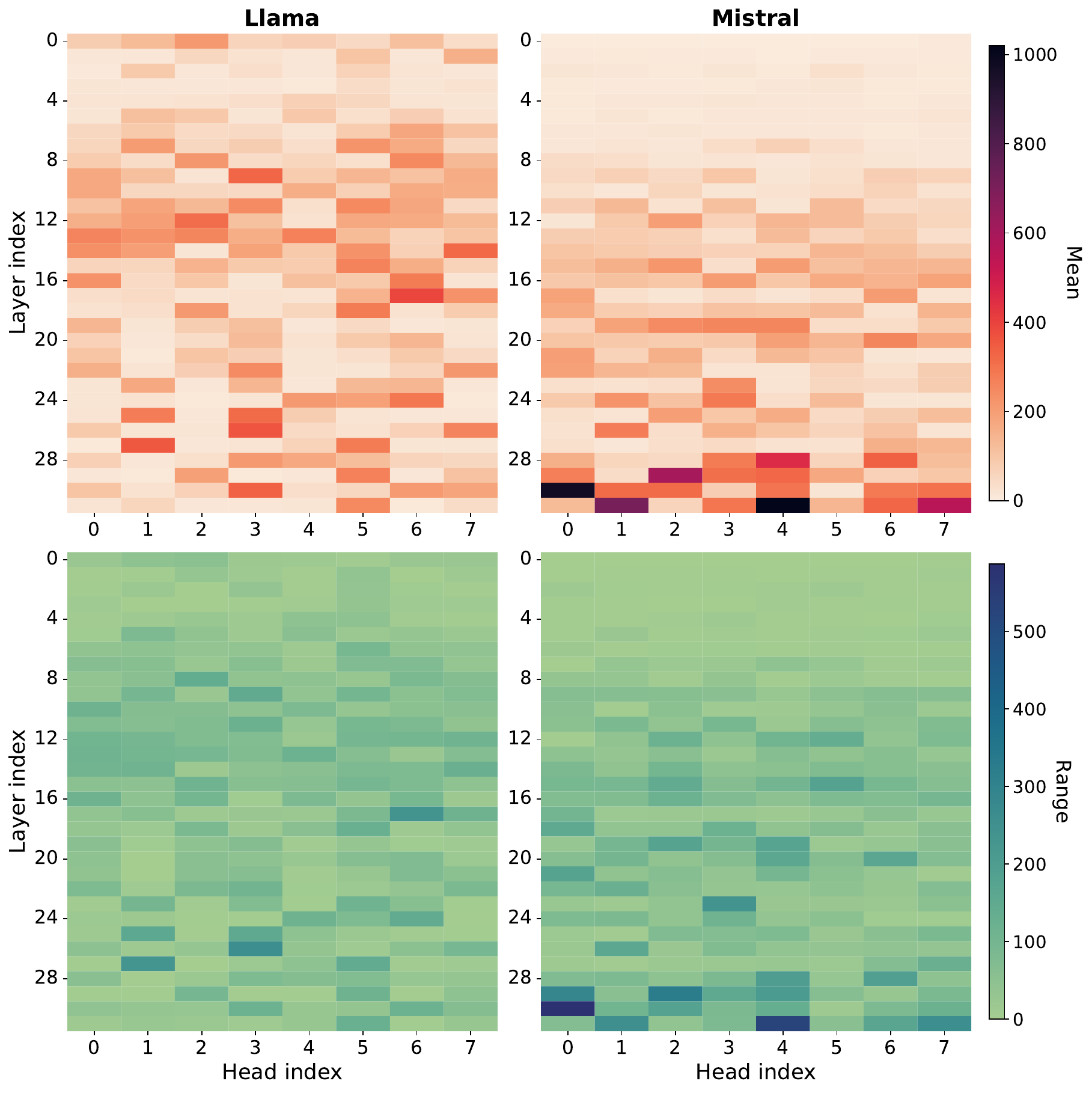}
  \caption{Retained tokens and variation per head}
  \label{fig:budget}
  \vspace{-10pt} % Adjust bottom spacing
\end{wrapfigure}

As shown in Figure \ref{fig:budget}, the number of retained tokens varies substantially across layers and heads. For example, in Llama, layers 13-14 retain significantly more tokens than layers 3 and 24. Similarly, within layer 27, a single head retains the majority of tokens.

We also observe distinct patterns across models. Llama concentrates retained tokens in the middle layers, whereas Mistral progressively shifts importance toward later layers. Moreover, these patterns are not stable and can vary by up to 500 tokens per head across inputs.

These results show that token contributions are highly non-uniform across layers and heads, vary across models, and are strongly input-dependent. \textbf{This underscores the need of our model-wide eviction strategy,} which dynamically selects tokens without offline calibration, and highlights a key limitation of calibration-based prior methods \cite{cake, headkv, pyramidkv}, which assume stable optimal budgets across diverse inputs.

\section{Conclusion}
This paper revisits KV cache eviction for long-context LLM inference and addresses two fundamental oversights in current methods.
First, we identify that current heuristic approaches neglect the structure of the attention layers; by considering the attention layer formulation, we provide a more complete measure of token importance. 
Second, we demonstrate that traditional head-level eviction is inherently suboptimal, as token importance is more accurately captured at the model-level.
Based on these insights, we introduce LaProx that reformulates cache eviction as a global matrix approximation problem rather than a head-wise weight-averaging task. 
\textbf{Our work is the first to explore a unified model-level cache eviction strategy, exposing the weakness of the long-standing traditional head-based heuristic schemes.}
Extensive evaluations across 19 datasets from LongBench and NIAH benchmarks confirm that this approach consistently outperforms prior strategies. Furthermore, our analysis validates that our method achieves higher fidelity to full-cache attention outputs than standard heuristics. Ultimately, this work introduces a different, global point of view to the study of KV cache, offering a principled foundation for efficient long-context inference.

\section*{Acknowledgments}
This work was supported by the Institute of Information \& Communications Technology Planning \& Evaluation (IITP) grants funded by the Korea government (MSIT) (\mbox{IITP-2026-RS-2020-II201847}, Information Technology Research Center; \mbox{IITP-2026-RS-2023-00256472}, Graduate School of Artificial Intelligence Semiconductor).

% \section*{References}
\printbibliography
%%%%%%%%%%%%%%%%%%%%%%%%%%%%%%%%%%%%%%%%%%%%%%%%%%%%%%%%%%%%

\newpage
\appendix

% \section{Technical appendices and supplementary material}
\section{Implementation details}
\label{appendix:imp}
Methods with uniform allocation assign equal cache capacity to each attention head, whereas non-uniform strategies, including ours, adaptively distribute cache capacity across heads or layers while keeping the total memory budget fixed. All baseline hyperparameters follow their default official implementations. For SLLM \cite{streamingllm}, we retain four sink tokens and allocate the remaining budget to a sliding recent window. All other methods employ an observation window of 32 tokens to limit overhead and apply average pooling with a kernel size of 7 to mitigate information fragmentation. The summary of the baselines is shown in Table \ref{table-baseline}.

To ensure compatibility with Grouped Query Attention (GQA) \cite{Ainslie2023GQATG}, we follow standard practice in prior works by using the mean attention weight within each query group as the selection criterion. All experiments are accelerated using FlashAttention-2 \cite{flashattention2}. And consistent with prior works \cite{snapkv,adakv}, cache eviction is applied in the prefilling phase per layer.

\begin{table*}[ht]
  \caption{Summary of evaluated methods on eviction score.}
  \label{table-baseline}
  \begin{center}
    \begin{small}
      % \begin{sc}
        \resizebox{\textwidth}{!}{
        \begin{tabular}{lcccc}
          \toprule
          \textsc{Methods}   & \textsc{Evict. Score}  & \textsc{Head Alloc.}      & \textsc{Layer Alloc.} & \textsc{Hyper Param.}  \\
          \midrule
          FullKV & - & - & - & - \\
          SLLM   & N/A (Fixed-indices eviction) & - & - & 1 (sink tokens) \\
          SnapKV & Mean$(\boldsymbol{A}[:,i])$ & - & - & 0 \\
          CAKE   & Mean$(\boldsymbol{A}[:,i])$ + $\gamma$ Var$(\boldsymbol{A}[:,i])$ & - & Entropy$(\boldsymbol{A})^{1/\tau_1}$ $\cdot$ Var$(\boldsymbol{A})^{1/\tau_2}$ & 3 ($\gamma, \tau_1, \tau_2$) \\
          AdaKV  & Mean$(\boldsymbol{A}[:,i])$ & $\text{Top-}K$(raw head scores) & - & 1 (safeguard)        \\
          CriticalKV  & (Mean$(\boldsymbol{A}[:,i]) + \epsilon)$ 
          % $\cdot$ 
          $\|\boldsymbol{VWo}[i,:]\|$ & - & - & 2 (safeguard, $\epsilon$)        \\
          LaProx (ours) & $\|\boldsymbol{A}[:,i]\|_2 \ \|\boldsymbol{VWo}[i,:]\|_2$ & $\text{Top-}K$(normalized model scores) & $\text{Top-}K$(normalized model scores) & 0 \\
          \bottomrule
        \end{tabular}
        }
      % \end{sc}
    \end{small}
  \end{center}
  % \vskip -0.1in
\end{table*}

\section{Additional Related Works on KV Cache Management}
\paragraph{KV Cache Eviction Metrics.} Beyond standard attention weights, several works explore alternative definitions of token importance. The work in \cite{l2evict} utilizes the $L2$ norm of key states to reduce computational overhead at the expense of accuracy. Other approaches, such as \cite{ahakv}, enhance attention-based metrics by incorporating value representations. However, these methods rely solely on value states, which only partially capture a token's contribution to the final representation. In practice, the output is further modulated by the output projection matrix ($W_O$)—a factor our ablation study (Section \ref{sec:ablation}) identifies as a critical.

CriticalKV \cite{criticalkv} is the most closely related to our work, as it also incorporates both value states and output projections. However, a fundamental conceptual gap remains: CriticalKV acts as a helper for other works and it treats output information merely as a scaling factor applied to the average-based eviction score. 
In contrast, \textbf{we reformulate KV cache eviction as a principled approximation of the attention matrix product} and \textbf{provide a complete solution}. By explicitly modeling the interaction between attention weights and projected value states as a unified operation, our method provides a more accurate estimation of token influence, as validated in Section \ref{sec:experiments}.

Furthermore, because CriticalKV lacks this rigorous matrix-product formulation, \textbf{it must rely on 2 hyper-parameters to maintain model performance} (an empirical safeguard and an $\epsilon$ to mitigate information loss). Its scoring also remains strictly head-centric, failing to address eviction as a layer-level optimization problem. Our approach moves beyond simple "scaling helpers" by providing a unified, theoretical-bound layer-wise solution that naturally handles budget allocation across heads without requiring empirical tuning.

\paragraph{KV Cache Selection.} 
While KV cache eviction methods reduce memory usage by retaining only a small subset of critical key–value entries, sparse attention approaches such as Quest \cite{quest} and InfiniGen \cite{infinigen} preserve the entire KV cache during inference but restrict computation to a selected subset of entries at each attention step. Or, some works \cite{duo,razor} combine the Dynamic Selection with Eviction approach by classifying attention heads into Streaming Heads with a limited Cache size and Retrieval Heads with a full Cache size.
By limiting attention computation rather than storage, sparse attention methods can significantly accelerate inference and often achieve high output quality. However, because all KV entries are still stored, these methods \textbf{do not reduce the memory footprint of the KV cache}.

\paragraph{KV Cache Quantization.} 
KV cache quantization methods reduce memory and computation costs by representing cached values using lower-precision formats. These approaches can be broadly divided into fixed-precision quantization, where all tokens share the same bit-width \cite{pqcache, flexgen}, and mixed-precision schemes, which allocate different bit-widths to different tokens \cite{kvquant, kivi}. 
However, while quantization effectively lowers per-token storage cost, the total \textbf{KV cache size continues to grow linearly with context length}, limiting its ability to address memory bottlenecks in very long-context settings.

% \section{Comparison and Compatibility with Quantization Methods}

\section{Details of Evaluation Benchmarks}
\label{appendix:benchmark}
\paragraph{LongBench Benchmark.} LongBench \cite{longbench} is a widely used long-context benchmark, serving as a standardized evaluation protocol commonly adopted by prior works and baselines \cite{snapkv,cake,adakv}. It consists of 16 datasets across 6 task domains: Single-Doc QA \cite{narrativeqa,qasper}, Multi-Doc QA \cite{hotpotqa,multihopqa,musique}, Summarization \cite{govreport,qmsum,multinews}, Few-shot Learning \cite{trec,triviaqa,samsum}, Synthetic Task \cite{passageretrieval}, and Code Completion \cite{Guo2023LongCoderAL,repobench}. The average token length across all 16 datasets is 6,711. More details can be found in Table \ref{tab:longbench}.

\begin{table}[tb]
  \caption{Details of each dataset in LongBench.}
  \label{tab:longbench}
  \centering
  \begin{small}
  \resizebox{\textwidth}{!}{
    \begin{tabular}{l l l r c c c} % Added one 'l' for the rotated label
    \toprule
    Task & Dataset & Source & Avg len & Eval Metric & Language & \#data\\
    \midrule
    
    \multirow{3}{*}{Single-Doc} 
    & NarrativeQA & Literature, Film & 18,409 & F1 & English & 200 \\
    & Qasper & Science & 3,619 & F1 & English & 200 \\
    & MultiFieldQA-en & Multi-field & 6,701 & F1 & English & 150 \\
    \midrule

    \multirow{3}{*}{Multi-Doc} 
    & HotpotQA & Wikipedia & 9,151 & F1 & English & 200 \\
    & 2WikiMultihopQA & Wikipedia & 4,887 & F1 & English & 200 \\
    & MuSiQue & Wikipedia & 11,214 & F1 & English & 200 \\
    \midrule

    \multirow{3}{*}{Summarization}
    & GovReport & Government report & 8,734 & Rouge-L & English & 200 \\
    & QMSum & Meeting & 10,614 & Rouge-L & English & 200 \\
    & MultiNews & News & 2,113 & Rouge-L & English & 200 \\
    \midrule

    \multirow{3}{*}{Few-shot}
    & TREC & Web question & 5,177 & Accuracy & English & 200 \\
    & TriviaQA & Wikipedia, Web & 8,209 & F1 & English & 200 \\
    & SAMSum & Dialogue & 6,258 & Rouge-L & English & 200 \\
    \midrule

    \multirow{2}{*}{Synthetic}
    & PassageCount & Wikipedia & 11,141 & Accuracy & English & 200 \\
    & PassageRetrieval-en & Wikipedia & 9,289 & Accuracy & English & 200 \\
    \midrule

    \multirow{2}{*}{Code}
    & LCC & Github & 1,235 & Edit Sim & Python/C\#/Java & 500 \\
    & RepoBench-P & Github & 4,206 & Edit Sim & Python/Java & 500 \\
    
    \bottomrule
    \end{tabular}
    }
  \end{small}
  \vskip -0.1in
\end{table}

\paragraph{Ruler Benchmark.}
Ruler \cite{ruler} is a diagnostic benchmark designed to evaluate long-context capabilities beyond simple retrieval. It consists of 13 datasets across 4 task domains: 
\begin{itemize}
    \item Retrieval: An extension of NIAH \cite{needle} that tests retrieval robustness using diverse needle types and varying quantities of hidden information.
    \item Multi-hop Tracing: Evaluates the model’s ability to track variable assignments and identify co-occurrence patterns that require connecting multiple pieces of information across the sequence.
    \item Aggregation: Tests the ability to identify the most frequent or common words distributed throughout the text.
    \item Question Answering: Tests the capability to answer questions where the answer is deeply embedded within extensive distracting or irrelevant content.
\end{itemize}
Further evaluations on more tasks from Ruler are provided in Appendix \ref{appendix:agnostic} and \ref{appendix:moe}.

\section{Ablation Study}
\label{appendix:ablation}
\subsection{Eviction Criteria Analysis}
\label{sec:ablation}
\begin{wrapfigure}{r}{0.5\textwidth}
% \vspace{-5pt}
  \begin{minipage}{\linewidth}
    \centering
    % First Table
    \captionof{table}{Comparison with different Eviction Indicators.}
    \label{tab:indicator}
    \begin{small}
      \begin{sc}
        \begin{tabular}{c c c}
        \toprule
        \multicolumn{2}{c}{\textbf{Eviction Indicator}} & \\
        Value State & Output Weight & Avg. \\
        \midrule
                   &            & 40.53 \\
        \checkmark &            & 41.69 \\
        \checkmark & \checkmark & 41.84 \\
        \bottomrule
        \end{tabular}
      \end{sc}
    \end{small}
    
    % Second Table
    \captionof{table}{Comparison with different Allocation Strategies.}
    \label{tab:alloc}
    \begin{small}
      \begin{sc}
      \resizebox{\linewidth}{!}{
        \begin{tabular}{c c c c}
        \toprule
        \multicolumn{3}{c}{\textbf{Allocation strategy}} & \\
        Head-Flatten & Layer-Flatten & Normalize & Avg. \\
        \midrule
                   &            &            & 41.84 \\
        \checkmark &            &            & 42.96 \\
        \checkmark & \checkmark &            & 41.14 \\
        \checkmark & \checkmark & \checkmark & 44.00 \\
        \bottomrule
        \end{tabular}}
      \end{sc}
    \end{small}
  \end{minipage}
  \vspace{-30pt} % Adjust bottom spacing to fit text wrap
\end{wrapfigure}

In this section, we present a series of ablation studies to evaluate the effectiveness of our proposed eviction strategy. We use Mistral-7B-Instruct-v0.3 with cache budget $B_{total} = 128L$ on LongBench as the default settings. 

\subsubsection*{Effectiveness of Eviction Indicator} To validate the necessity of our metric, we isolate the influence of $V$ from $VW_O$ then compare them against the baseline. As shown in Table \ref{tab:indicator}, incorporating $V$ improves performance over attention-only baselines, but remains sub-optimal as it still provides an incomplete computation of the attention output. The highest accuracy is achieved by our full $AVW_O$ indicator, demonstrating that the output projection is essential for quantifying a token's contribution to the layer's output.

\subsubsection*{Effectiveness of Allocation Strategy} We further analyze the necessity of each component in our allocation method, specifically flattening (head-wise and layer-wise) and normalization (Table \ref{tab:alloc}). 
While head-wise allocation provides immediate performance gains, layer-wise allocation degrades performance to below the baseline when the scores are used in their raw form. 
This decline stems from raw score scale differences across layers; without normalization, the budget focuses only on high-magnitude layers while starving others. 
Despite its simplicity, our normalization scheme resolves this bias, allowing layer-wise allocation to effectively complement head-wise settings and further improve performance.

% \textbf{$VW_O$ projection and Allocation Dependency.} 
While the allocation gain shown in Table \ref{tab:alloc} is more noticeable, it is worth noting that our method still outperforms prior approaches without it.
Furthermore, \textbf{projection and allocation strategy are not independent; rather, the former is the mathematical prerequisite for the latter.} This is because without our projection, attention-based score is limited in its own head and not comparable globally.

To validate this, we experiment with 4 different settings: Uniform budget and Global allocation with Attention score ($A_L$ and $A_G$), Uniform budget and Global allocation with LaProx score ($L_L$ and $L_G$). The experiments were conducted using LongBench benchmark under the budget of 128 tokens. The results are shown in Table \ref{tab:score-alloc}.

\begin{table*}[htb]
\caption{Comparison across 16 LongBench datasets for the cache budget of 128L. The best result is highlighted in \textbf{bold}.}
\label{tab:score-alloc}
\centering
\small
\setlength{\tabcolsep}{3pt}
\resizebox{\textwidth}{!}{
\begin{tabular}{l ccc ccc ccc ccc cc cc c}
\toprule
\multirow{2}{*}{Method} & \multicolumn{3}{c}{Single-Document QA} & \multicolumn{3}{c}{Multi-Document QA} & \multicolumn{3}{c}{Summarization} & \multicolumn{3}{c}{Few-shot Learning} & \multicolumn{2}{c}{Synthetic} & \multicolumn{2}{c}{Code} & \multirow{2}{*}{Avg.} \\
\cmidrule(lr){2-4} \cmidrule(lr){5-7} \cmidrule(lr){8-10} \cmidrule(lr){11-13} \cmidrule(lr){14-15} \cmidrule(lr){16-17}

 & NrtvQA & Qasper & MF-en & HotpotQA & 2WikiMQA & Musique & GovRep & QMSum & MultiNews & TREC & TriviaQA & SAMSum & PCount & PR-en & Lcc & RB-P &  \\

\midrule
\multicolumn{18}{c}{Meta-Llama-3.1-8B-Instruct, $B_{\text{total}} = 128L$} \\
\midrule
% Full & 30.28 & 45.46 & 54.94 & 56.02 & 46.5 & 31.28 & 35.11 & 25.28 & 27.19 & 72.5 & 91.65 & 43.86 & 8.88 & 99.5 & 65.13 & 58.7 & 49.51 \\
$A_L$ & 25.53 & 23.92 & 48.15 & 52.94 & 43.5 & 26.67 & 19.48 & 22.47 & 19.68 & 47 & 89.56 & 39.51 & 8.41 & 98 & 59.43 & 52.84 & 42.31 \\
$A_G$ & 24.98 & 27.16 & 49.96 & 52.1 & 43.38 & 27.73 & 21.21 & 23.21 & 19.93 & 48.5 & \textbf{89.95} & 40.1 & 7.92 & 99 & 60.45 & 53.24 & 43.05 \\
$L_L$ & 25.49 & 26.72 & 49.91 & 54.7 & 42.81 & 29.18 & 20.67 & \textbf{23.67} & 19.72 & 48 & 89.36 & \textbf{41.23} & 8.88 & 99 & 59.18 & 53.49 & 43.26 \\
$L_G$ & \textbf{25.78} & \textbf{31.38} & \textbf{52.97} & \textbf{55.17} & \textbf{44.95} & \textbf{29.46} & \textbf{21.66} & 23.66 & \textbf{21.47} & \textbf{59} & 89.94 & 41.09 & \textbf{8.89} & \textbf{99.5} & \textbf{61.91} & \textbf{56.16} & \textbf{45.19} \\
% \midrule
% \multicolumn{18}{c}{Mistral-7B-Instruct-v0.3, $B_{\text{total}} = 128L$} \\
% \midrule
% % Full    & 29.07 & 41.58 & 52.88 & 49.37 & 39.01 & 28.58 & 34.81 & 25.66 & 27.82 & 76 & 88.59 & 47.4 & 5.5 & 98 & 61.4 & 62.53 & 48.01 \\
% $A_L$   & 23.81 & 25.05 & 46.92 & 44.98 & 35.06 & 23.11 & 20.49 & 21.55 & 19.32 & 43.5 & \textbf{89.78} & 42.87 & 6 & 93.5 & 55.95 & 54.6 & 40.41 \\
% $A_G$   & 26.62 & 27.63 & 51.67 & 47.63 & 36.51 & 26.14 & 20.85 & 23.34 & 21.97 & 55.5 & 88.66 & 43.6 & 4 & 96 & 50.38 & 55.23 & 42.23 \\
% $L_L$   & 25.33 & 27.42 & 49.29 & 49.09 & 35.58 & 24.32 & 21.14 & 22.4 & 21.27 & 52 & 89.01 & 43.23 & 6.5 & 93 & 55.47 & 54.32 & 41.84 \\
% $L_G$   & \textbf{28} & \textbf{30.26} & \textbf{54.02} & \textbf{47.86} & \textbf{37.54} & \textbf{25.54} & \underline{21.26} & \textbf{23.31} & \textbf{22.1} & \textbf{63.5} & \underline{89.7} & \underline{43.46} & 5 & \textbf{96} & \textbf{58.85} & \textbf{57.66} & \textbf{44.00} \\

\bottomrule
\end{tabular}
}
\vskip -0.1in
\end{table*}

As seen from the results, \textbf{without $VW_O$, global allocation offers no improvement compared to the baseline.} Particularly, the performance of $A_G$ is similar to $A_L$, and is lower than $L_L$, which is our method without any allocation.

\subsubsection*{Moving from sampling to deterministic} The Monte Carlo matrix multiplication framework derives the optimal importance measure for each row/column by showing that the sampling probability
\[
p_i \propto
\left\lVert A[:,i] \right\rVert_2
\left\lVert B[i,:] \right\rVert_2
\]
minimizes the expected Frobenius-norm approximation error. Our work adopts this theoretically grounded importance measure but replaces stochastic sampling with deterministic Top-K selection.

\textbf{The Monte Carlo framework minimizes the expected (i.e., average over repeated trials) approximation error} over the sampling distribution. However, \textbf{LLM inference executes a single-shot forward pass}.
Consequently, stochastic sampling may discard high-contribution KV pairs due to sampling randomness, introducing unnecessary approximation error in that particular inference. Deterministic Top-K instead retains the highest-scoring KV pairs according to the same theoretically derived importance measure, \textbf{eliminating sampling variance} while preserving the intended ranking.

To validate this design choice, we additionally compare stochastic sampling (``Sample'') and deterministic Top-K (``Deter.'') using the identical norm-product score on the LongBench benchmark with the Mistral-7B-Instruct-v0.3 model under a 128-token budget.

\begin{table}[ht]
    \centering
    \caption{Comparison between deterministic Top-K selection and stochastic
    sampling using the same norm-product importance score on LongBench.}
    \label{tab:det_vs_sample}
    \resizebox{\linewidth}{!}{%
    \begin{tabular}{lccccccccccccccccc}
        \toprule
        Method & NrtvQA & Qasper & MF & Hotpot & 2WikiQA & MuSiQue & GovRep & QMSum & MultNews & TREC & TriviaQA & SAMSum & PCount & PR & LCC & RB-P & Avg. \\
        \midrule

        Deter. & \textbf{28.00} & \textbf{30.26} & \textbf{54.02} & \textbf{47.86} & \textbf{37.54} & \textbf{25.54} & \textbf{21.26} & \textbf{23.31} & \textbf{22.10} & \textbf{63.50} & 89.70 & \textbf{43.46} & 5.00 & \textbf{96.00} & \textbf{58.85} & \textbf{57.66} & \textbf{44.00} \\

        SnapKV & 23.81 & 25.05 & 46.92 & 44.98 & 35.06 & 23.11 & 20.49 & 21.55 & 19.32 & 43.50 & \textbf{89.78} & 42.87 & \textbf{6.00} & 93.50 & 55.95 & 54.60 & 40.41 \\

        Sample & 22.73 & 24.61 & 41.54 & 39.73 & 30.66 & 16.38 & 14.99 & 21.94 & 18.58 & 55.00 & 83.10 & 39.59 & 5.00 & 94.67 & 37.16 & 38.45 & 36.48 \\

        \bottomrule
    \end{tabular}%
    }
\end{table}

As shown in Table~\ref{tab:det_vs_sample},
\textbf{stochastic sampling significantly degrades performance}, even below
the SnapKV baseline, despite using the same theoretically derived importance
score. In contrast, deterministic Top-K (LaProx) consistently obtains the
highest scores across the majority of datasets and achieves the best overall
performance.

\subsection{Sensitivity Analysis}
\label{appendix:ablation-sensitivity}

\textbf{Window size robustness.} Following the configurations of the baselines \cite{snapkv,cake,adakv}, we utilized a default historical window size of 32 for our primary experiments. We note that the \emph{hyperparameter-free} property of LaProx refers specifically to its eviction scoring and budget allocation mechanisms, which require no method-specific tunable coefficients. The historical window, by contrast, is a common structural component of KV-cache compression frameworks used to preserve local context and reduce attention-computation overhead.

To assess the sensitivity of our method to this hyperparameter, we conducted an ablation study on the Mistral-7B-Instruct-v0.3 model (with a cache budget of 128) across four window sizes: 8, 16, 32, and 64. The results are summarized in Table \ref{tab:window}.
LaProx maintains stable performance across window sizes 8 through 32, with scores ranging between 43.79 and 44.10. However, performance degrades at a window size of 64; this occurs because the large observation window reduces the candidate space for eviction, thereby limiting the flexibility of our eviction strategy. Overall, these results demonstrate that LaProx is highly robust to variations in the historical window size.

\begin{table*}[ht]
\caption{Analysis on different observation window sizes.}
\label{tab:window}
\centering
\small
\setlength{\tabcolsep}{3pt}
\resizebox{\textwidth}{!}{
\begin{tabular}{l ccc ccc ccc ccc cc cc c}
\toprule
\multirow{2}{*}{Method} & \multicolumn{3}{c}{Single-Document QA} & \multicolumn{3}{c}{Multi-Document QA} & \multicolumn{3}{c}{Summarization} & \multicolumn{3}{c}{Few-shot Learning} & \multicolumn{2}{c}{Synthetic} & \multicolumn{2}{c}{Code} & \multirow{2}{*}{Avg.} \\
\cmidrule(lr){2-4} \cmidrule(lr){5-7} \cmidrule(lr){8-10} \cmidrule(lr){11-13} \cmidrule(lr){14-15} \cmidrule(lr){16-17}
 & NrtvQA & Qasper & MF-en & HotpotQA & 2WikiMQA & Musique & GovRep & QMSum & MultiNews & TREC & TriviaQA & SAMSum & PCount & PR-en & Lcc & RB-P &  \\

\midrule
\multicolumn{18}{c}{Mistral-7B-Instruct-v0.3, $B_{\text{total}} = Full$} \\
\midrule
FullKV & 29.07 & 41.58 & 52.88 & 49.37 & 39.01 & 28.58 & 34.81 & 25.66 & 27.82 & 76 & 88.59 & 47.4 & 5.5 & 98 & 61.4 & 62.53 & 48.01 \\
\midrule
\multicolumn{18}{c}{Mistral-7B-Instruct-v0.3, $B_{\text{total}} = 128L$} \\
\midrule
w=8   & 27.87 & 30.51 & 53.85 & 48.2 & 35.95 & 25.98 & 22.58 & 23.2 & 22.24 & 72 & 88.66 & 43.68 & 5.5 & 94 & 56.99 & 54.4 & 44.10 \\
w=16  & 25.75 & 31.43 & 52.3 & 47.46 & 35.57 & 25.55 & 22.93 & 22.98 & 22.26 & 69.5 & 88.44 & 44.17 & 5.5 & 94 & 57.43 & 55.34 & 43.79 \\
w=32  & 28 & 30.26 & 54.02 & 47.86 & 37.54 & 25.54 & 21.26 & 23.31 & 22.1 & 63.5 & 89.7 & 43.46 & 5 & 96 & 58.85 & 57.66 & 44.00 \\
w=64  & 23.24 & 24.99 & 45.7 & 48.43 & 35.28 & 24.29 & 19.13 & 21.2 & 19.9 & 52 & 88.74 & 43.28 & 3.5 & 92 & 55.53 & 55.08 & 40.77 \\

\bottomrule
\end{tabular}
}
\vskip -0.1in
\end{table*}

\subsection{Efficiency Analysis}
\label{appendix:efficiency}
\subsubsection*{Latency Breakdown}
Since our method differs from the baselines primarily in the eviction score computation and allocation stages (while all other components, such as attention computation and linear projections, remain identical), we focus our analysis on these components.

To better understand the source of latency overhead, we focus on the attention layer and decompose its forward pass into four stages: attention computation (with a historical window), eviction score computation, allocation, and final output projection. We then profile each stage separately.
All measurements are taken at the last layer at the end of the prefill stage, where KV cache eviction is performed. For other layers, eviction is not triggered and the remaining computations are identical. The results are summarized in the table below.

\begin{table}[htb]
  \caption{Latency breakdown (s).}
  \begin{center}
    \begin{small}
        \begin{tabular}{l c c c c c c}
        \toprule
         & Attn Comp & Evict. Comp & Allocation & Proj & Total \\
        \midrule
        % CAKE   & --  & --  & --  & --  & --  \\
        AdaKV  & 0.03898  & 0.00042  & 0.10230 & 0.62226  & 0.76396  \\
        LaProx & 0.03889  & 0.04174  & 0.09684  & 0.6162  & 0.79368  \\
        \bottomrule
        \end{tabular}
    \end{small}
  \end{center}
\end{table}

The results show that the latency of attention computation and output projection remains unchanged, as these components share the same implementation across methods. The allocation stage is slightly faster in LaProx, since it does not require additional safeguard logic (for example, threshold checks and budget adjustments to enforce constraints).
In contrast, the eviction score computation in LaProx introduces higher latency compared to AdaKV. However, \textbf{when placed in the context of the full pipeline, this overhead is minimal,} as other components (e.g., linear projections) dominate the overall latency.

% For ease of implementation, our current eviction score is computed using standard PyTorch operators. While functional, this approach can be further optimized by fusing the $V \times W_O$ multiplication, squaring, and reduction steps into a single CUDA or Triton kernel.

\subsubsection*{Scalability analysis} In Section \ref{sec:efficiency}, we have evaluated the efficiency of LaProx on context lengths up to 128K using Meta-Llama-3.1-8B-Instruct, which employs 32 attention heads with GQA. These experiments demonstrate that the proposed scoring mechanism maintains low overhead even under extremely long contexts and modern multi-head architectures.

To further strengthen this point, we additionally evaluate scalability with respect to batch size. The experiments are conducted on Meta-Llama-3.1-8B-Instruct with a 64K context length and a cache budget of 128 tokens, while comparing against the two strongest baselines, AdaKV and CriticalKV.

\begin{table}[t]
    \centering

    % ---------------- Left table ----------------
    \begin{minipage}[t]{0.49\linewidth}
        \centering
        \caption{Prefilling latency (seconds).}
        \label{tab:batch_prefill}

        \resizebox{\linewidth}{!}{%
        \begin{tabular}{lcccc}
            \toprule
            Batch size & FullKV & AdaKV & CriticalKV & LaProx \\
            \midrule
            1 & 8.586  & 9.854  & 10.471 & 10.619 \\
            2 & 16.769 & 19.803 & 20.263 & 20.728 \\
            3 & OOM    & 28.398 & 29.187 & 30.025 \\
            \bottomrule
        \end{tabular}%
        }
    \end{minipage}
    \hfill
    % ---------------- Right table ----------------
    \begin{minipage}[t]{0.49\linewidth}
        \centering
        \caption{Decoding latency (seconds).}
        \label{tab:batch_decode}

        \resizebox{\linewidth}{!}{%
        \begin{tabular}{lcccc}
            \toprule
            Batch size & FullKV & AdaKV & CriticalKV & LaProx \\
            \midrule
            1 & 0.031 & 0.027 & 0.026 & 0.025 \\
            2 & 0.043 & 0.030 & 0.025 & 0.031 \\
            3 & OOM   & 0.026 & 0.027 & 0.027 \\
            \bottomrule
        \end{tabular}%
        }
    \end{minipage}

\end{table}

As shown in Tables~\ref{tab:batch_prefill} and~\ref{tab:batch_decode}, all KV eviction methods scale similarly with increasing batch size. 
Compared with AdaKV and CriticalKV, LaProx introduces only a marginal increase in prefilling latency while maintaining comparable decoding latency. 
More importantly, all eviction-based methods can support the configuration where the FullKV baseline runs out of memory (batch size $=3$). 
These results indicate that the additional Frobenius-norm computations do not become the dominant runtime bottleneck, even for long contexts, multiple attention heads, or large batch sizes.

\section{Limitations}
\label{limitation}
Our work is the first to introduce a unified eviction framework that assigns globally comparable importance scores to tokens, enabling model-wide joint selection of KV cache entries. To achieve practical efficiency, our current formulation relies on relatively simple approximation and normalization schemes, which we view as an initial foundation for future model-wide KV cache research.
Future work may explore more sophisticated matrix multiplication approximation schemes or different normalization methods for global token comparison to further improve cache eviction performance.

\section{Proofs}
\subsection{Proof of Remark \ref{rem:decompose}}
\label{proof:MHA-decomposition}

Let:
\begin{itemize}
    \item $h$ be the number of attention heads.
    \item $d_v$ be the dimension of each head.
    \item $d_{model}$ be the model dimension.
    \item $H^i \in \mathbb{R}^{n \times d_v}$ be the output of the $i$-th head ($H^i = \text{Attention}(Q^i, K^i, V^i)$).
    \item $W_O \in \mathbb{R}^{(h \cdot d_v) \times d_{model}}$ be the output weight matrix.
\end{itemize}

Then, the standard MHA is defined as the concatenation of all head outputs followed by a linear projection:
\begin{equation}
    \text{Output} = \text{Concat}(H^1, H^2, \dots, H^h) W_O
\end{equation}

We can represent the concatenation as a partitioned row matrix:
\begin{equation}
    H = \begin{bmatrix} H^1 & H^2 & \dots & H^h \end{bmatrix}
\end{equation}

We can similarly partition the weight matrix $W_O$ vertically into $h$ blocks, where each $W_O^i \in \mathbb{R}^{d_v \times d_{model}}$:
\begin{equation}
    W_O = \begin{bmatrix} W_O^1 \\ W_O^2 \\ \vdots \\ W_O^h \end{bmatrix}
\end{equation}

By applying the rules of block matrix multiplication:
\begin{equation}
\begin{aligned}
    \text{Output} &= H W_O \\
    &= \begin{bmatrix} H^1 & H^2 & \dots & H^h \end{bmatrix} \begin{bmatrix} W_O^1 \\ W_O^2 \\ \vdots \\ W_O^h \end{bmatrix} \\
    &= (H^1 W_O^1) + (H^2 W_O^2) + \dots + (H^h W_O^h) \\
    &= \sum_{i=1}^{h} H^i W_O^i
\end{aligned}
\end{equation} \hfill$\square$

\subsection{Proof of Remark \ref{rem:layer_token}}
\label{appendix:layer-importance}
% \textit{Theorem.} For a multi-head attention layer $l$, the contribution of token $j$ at query position $i$ is
% \begin{equation}
% \Delta \mathbf{o}^l(i,j)
% =
% \sum_{h=1}^{H}
% A^{l,h}(i,j)\,
% V^{l,h}(j)\,
% W_{O}^{l,h}
% \label{eq:token_contribution1}
% \end{equation}
% and the layer output satisfies
% \begin{equation}
% \mathbf{o}^l(i)
% =
% \sum_{j}
% \Delta \mathbf{o}^l(i,j).
% \label{eq:layer_output}
% \end{equation}
% Therefore, the contribution of a token from any head enters the layer output directly through linear addition, and removing a token from a single head induces an immediate change in the layer output. Token importance must therefore be evaluated at the layer level.

% \textit{Proof.} 
Consider a transformer layer $l$ with $H$ attention heads. Let $i$ denote the current query position and $j$ a cached token position. The output of multi-head attention at layer $l$ is
\begin{equation}
\mathbf{o}^l(i)
=
\sum_{h=1}^{H}
\mathbf{o}^{l,h}(i)\, W_{O}^{l,h},
\end{equation}
where the output of head $h$ is
\begin{equation}
\mathbf{o}^{l,h}(i)
=
\sum_{j}
A^{l,h}(i,j)\, V^{l,h}(j).
\end{equation}
Substituting and reordering terms yields
\begin{equation}
\mathbf{o}^l(i)
=
\sum_{j}
\sum_{h=1}^{H}
A^{l,h}(i,j)\,
V^{l,h}(j)\,
W_{O}^{l,h}.
\label{eq:token_decomposition}
\end{equation}

Equation~\eqref{eq:token_decomposition} admits a natural decomposition over cached tokens. In particular, the contribution of token $j$ to the layer output can be written as
\begin{equation}
\Delta \mathbf{o}^l(i,j)
=
\sum_{h=1}^{H}
A^{l,h}(i,j)\,
V^{l,h}(j)\,
W_{O}^{l,h}.
\label{eq:token_contribution}
\end{equation}
This expression shows that a token’s effect on the layer output is not localized to any single head, but instead accumulates additively across all heads without any nonlinear operations. \hfill$\square$

% Suppose one attempts to evict token $j$ from only a subset of heads $\mathcal{H}' \subset \{1,\dots,H\}$. The resulting layer output becomes
% \begin{equation}
% \tilde{\mathbf{o}}^l(i)
% =
% \mathbf{o}^l(i)
% -
% \sum_{h \in \mathcal{H}'}
% A^{l,h}(i,j)\,
% V^{l,h}(j)\,
% W_{O}^{l,h}.
% \end{equation}

% The output perturbation caused by removing token $j$ from the layer is measured by
% \begin{equation}
% \left\|
% \Delta \mathbf{o}^l(i,j)
% \right\|
% =
% \left\|
% \sum_{h}
% A^{l,h}(i,j)\,
% V^{l,h}(j)\,
% W_{O}^{l,h}
% \right\|
% \;\ge\;
% \max_h
% \left\|
% A^{l,h}(i,j)\,
% V^{l,h}(j)\,
% W_{O}^{l,h}
% \right\|.
% \label{eq:norm_lower_bound}
% \end{equation}
% This lower bound implies that a token exhibiting large influence in \emph{any} head necessarily induces a non-negligible perturbation at the layer output.

\section{Extended Experimental Results}

\subsection{Extended Evaluation on InfiniteBench Benchmark}
\label{sec:infinitebench}
To further assess the performance of our proposal under extreme long-context settings, we conducted additional evaluations on the InfiniteBench benchmark \cite{infinitebench} and compared against 2 representative baselines, AdaKV and CriticalKV.

Since InfiniteBench involves extremely long contexts (including some datasets with an average input length of upto 190K+ tokens), we evaluate on Meta-Llama-3.1-8B-Instruct, which supports 128K context lengths. To balance evaluation cost, we set the cache budget to 1024 tokens and the dataset limit to 100 samples per dataset. The results are summarized in Table \ref{tab:infinitebench}.
\begin{table}[htb]
\caption{Comparison of different methods on InfiniteBench. Best result is highlighted in \textbf{bold}.}
\centering
\small
\setlength{\tabcolsep}{3pt}
\label{tab:infinitebench}
\resizebox{\columnwidth}{!}{
\begin{tabular}{lccccccccccc}
\toprule
Method & En.Sum & En.QA & En.MC & En.Dia & Code.Debug & Code.Run & Math.Find & Ret.Pass & Ret.Num & Ret.KV & Avg. \\
\midrule
\multicolumn{11}{c}{Meta-Llama-3.1-8B-Instruct, $B_{\text{total}} = 1024L$} \\
\midrule
AdaKV       & 23.32 & 8.66 & \textbf{71} & 9  & \textbf{22} & 1 & \textbf{53} & \textbf{100} & 94 & 2 & 38.40 \\
CriticalKV  & \textbf{23.34} & 9.31 & 70 & 8  & \textbf{22} & \textbf{2} & \textbf{53} & \textbf{100} & 93 & 0 & 38.07 \\
LaProx      & 21.64 & \textbf{9.83} & \textbf{71} & \textbf{10} & \textbf{22} & \textbf{2} & \textbf{53} & \textbf{100} & \textbf{96} & \textbf{4} & \textbf{38.95} \\
% \midrule
% \multicolumn{11}{c}{Qwen2.5-7B-Instruct, $B_{\text{total}} = 1024L$} \\
% \midrule
% AdaKV       & 20.83 & \textbf{5.65} & \textbf{50} & 8 & \textbf{30} & 0 & 92 & 78 & 0 & 0 & 28.45 \\
% CriticalKV  & 20.67 & 5.32 & \textbf{50} & 6 & \textbf{30} & 0 & 70 & 70 & \textbf{4} & 0 & 25.60 \\
% LaProx      & \textbf{21.40} & 4.71 & \textbf{50} & \textbf{10} & \textbf{30} & 0 & \textbf{98} & \textbf{98} & \textbf{4} & 0 & \textbf{31.62} \\
\bottomrule
\end{tabular}
}
\end{table}

The results show that, on this highly challenging long-context benchmark, LaProx consistently achieves stronger performance than our strongest baseline across most datasets for both models. These results demonstrate the robustness of our method under extreme long-context settings, as it maintains strong performance across different models and evaluation tasks.

\subsection{Evaluations on Question-Agnostic Setting}
\label{appendix:agnostic}
In our main experiments, we follow the standard practice of the prior works \cite{snapkv,pyramidkv,cake}, where the question is compressed together with the context, to provide a unified evaluation scenario. In this section, we adopt a question-agnostic setting, where the context is compressed without any knowledge of the question, providing a more challenging evaluation. 
The results are summarized in Tables \ref{tab:agnostic-longbench}-\ref{tab:agnostic-ruler}.

Evaluating RULER \cite{ruler} is particularly computationally expensive, as it consists of 13 synthetic tasks and requires approximately 17 GPU hours for a single 32K-context evaluation per configuration. As a result, conducting a comprehensive sweep across all settings becomes prohibitively costly. Therefore, in this Appendix, we restrict comparisons to 2 representative baselines, AdaKV and CriticalKV.

\begin{table*}[ht]
\caption{Comparison across 16 LongBench datasets. The best result is highlighted in \textbf{bold}.}
\label{tab:agnostic-longbench}
\centering
\small
\setlength{\tabcolsep}{3pt}
\resizebox{\textwidth}{!}{
\begin{tabular}{l ccc ccc ccc ccc cc cc c}
\toprule

\multirow{2}{*}{Method} & \multicolumn{3}{c}{Single-Document QA} & \multicolumn{3}{c}{Multi-Document QA} & \multicolumn{3}{c}{Summarization} & \multicolumn{3}{c}{Few-shot Learning} & \multicolumn{2}{c}{Synthetic} & \multicolumn{2}{c}{Code} & \multirow{2}{*}{Avg.} \\
\cmidrule(lr){2-4} \cmidrule(lr){5-7} \cmidrule(lr){8-10} \cmidrule(lr){11-13} \cmidrule(lr){14-15} \cmidrule(lr){16-17}

 & NrtvQA & Qasper & MF-en & HotpotQA & 2WikiMQA & Musique & GovReport & QMSum & MultiNews & TREC & TriviaQA & SAMSum & PCount & PR-en & Lcc & RB-P &  \\

\midrule
\multicolumn{18}{c}{Meta-Llama-3.1-8B-Instruct, 20\% Cache} \\
\midrule
FullKV & 29.21 & 44.6 & 55.73 & 58.14 & 49.26 & 32.61 & 34.65 & 25.24 & 26.9 & 73 & 92.45 & 43.42 & 7.62 & 99.5 & 63.58 & 52.73 & 49.29 \\
% SnapKV  & 11.02 & 30.06 & 28.14 & 50.65 & 39.4 & 17.49 & 27.23 & 18.26 & 20.79 & 66 & 88.96 & 46.65 & 12 & 78.5 & 73.78 & 67.92 & 42.31 \\
AdaKV   & 27.02 & 29.01 & 33.12 & 49.19 & 29.45 & 21.24 & 26.86 & 21.84 & 22.83 & 54 & 91.33 & 43.84 & 6.12 & 80 & 66.59 & 55.38 & 41.11 \\
% CAKE    & 12.86 & 15.86 & 19.78 & 31.39 & 27.08 & 8.39 & 18.48 & 14.59 & 17.47 & 26 & 87.13 & 37.83 & 2 & 5 & 67.04 & 69.02 & 28.75 \\
CriticalKV    & 29.56 & 30.02 & 33.25 & 51.37 & 33.98 & 25.5 & 28.84 & 22.53 & 22.92 & 54 & 91.75 & 43.85 & \textbf{7.61} & 95.5 & 66.73 & 54.68 & 43.25 \\
LaProx    & \textbf{31.34} & \textbf{41.26} & \textbf{51.99} & \textbf{55.04} & \textbf{39.25} & \textbf{26.36} & \textbf{29.95} & \textbf{24.04} & \textbf{24.05} & \textbf{68} & \textbf{92.2} & \textbf{43.83} & 6.91 & \textbf{99.5} & \textbf{68.02} & \textbf{56.09} & \textbf{47.36} \\

\bottomrule
\end{tabular}
}
\vskip -0.1in
\end{table*}

\begin{table*}[ht]
\caption{ Comparison across 13 Ruler datasets. The best result is highlighted in \textbf{bold}.}
\centering
\small
\setlength{\tabcolsep}{3pt}
\label{tab:agnostic-ruler}

\resizebox{\textwidth}{!}{
\begin{tabular}{lcccccccccccccc}
\toprule
Method & CWE & FWE & NIAH\_MK1 & NIAH\_MK2 & NIAH\_MK3 & NIAH\_MQ & NIAH\_MV & NIAH\_S1 & NIAH\_S2 & NIAH\_S3 & QA1 & QA2 & VT & Avg. \\
\midrule
\multicolumn{15}{c}{Meta-Llama-3.1-8B-Instruct, 20\% Cache} \\
\midrule
FullKV & 51 & 94.67 & 100 & 100 & 100 & 97.5 & 99.5 & 100 & 100 & 100 & 84 & 54 & 100 & 90.82 \\
% SnapKV & 0.7 & 95.67 & 9 & 5 & 4 & 9 & 6 & 25 & 22 & 2 & 34 & 48 & 55.4 & 24.29 \\
AdaKV  & 10 & 85.33 & 82 & 21 & 25 & 74 & 62.5 & 98 & 94 & 21 & 34 & 47 & 92.2 & 57.38 \\
% CAKE   &  &  &  &  &  &  &  &  &  &  &  &  &  &  \\
CriticalKV   & \textbf{12.6} & 75.33 & 88 & 14 & 5 & 86.25 & 77.25 & 99 & \textbf{100} & 18 & 44 & 48 & 90 & 58.26 \\
LaProx   & 12 & \textbf{90.33} & \textbf{100} & \textbf{99} & \textbf{97} & \textbf{97.5} & \textbf{96.75} & \textbf{100} & \textbf{100} & \textbf{97} & \textbf{65} & \textbf{54} & \textbf{99.2} & \textbf{85.21} \\
\bottomrule
\end{tabular}
}
\vskip -0.1in
\end{table*}

The results in Table \ref{tab:agnostic-longbench} and \ref{tab:agnostic-ruler} reveal that, under this challenging setting, methods that do not leverage $VW_O$ information suffer significant performance degradation. In contrast, LaProx consistently maintains performance close to the full KV cache and outperforms all competing approaches. Furthermore, on the more challenging NIAH variants (multi-needle), most prior methods fail to reliably retrieve the needles, while LaProx maintains performance comparable to the FullKV setting.

\subsection{Evaluations on Mixture-of-Experts}
\label{appendix:moe}
To demonstrate that our approach easily adapts across architectures, we evaluate it on Mixture-of-Experts (MoE) models. Although MoE introduces sparsity in the Feed-Forward Networks (FFN), the attention layers—where the KV cache resides—remain unchanged. Since cache eviction operates within attention, our method transfers directly to MoE without any architectural modifications. 
In this experiment, we use Qwen3-30B-A3B-Instruct-2507 and perform the same tasks and settings with Appendix \ref{appendix:agnostic}. The results are presented in Tables \ref{tab:moe-longbench} and \ref{tab:moe-ruler}.

\begin{table*}[ht]
\caption{Comparison across 16 LongBench datasets. The best result is highlighted in \textbf{bold}.}
\label{tab:moe-longbench}
\centering
\small
\setlength{\tabcolsep}{3pt}
\resizebox{\textwidth}{!}{
\begin{tabular}{l ccc ccc ccc ccc cc cc c}
\toprule

\multirow{2}{*}{Method} & \multicolumn{3}{c}{Single-Document QA} & \multicolumn{3}{c}{Multi-Document QA} & \multicolumn{3}{c}{Summarization} & \multicolumn{3}{c}{Few-shot Learning} & \multicolumn{2}{c}{Synthetic} & \multicolumn{2}{c}{Code} & \multirow{2}{*}{Avg.} \\
\cmidrule(lr){2-4} \cmidrule(lr){5-7} \cmidrule(lr){8-10} \cmidrule(lr){11-13} \cmidrule(lr){14-15} \cmidrule(lr){16-17}

 & NrtvQA & Qasper & MF-en & HotpotQA & 2WikiMQA & Musique & GovReport & QMSum & MultiNews & TREC & TriviaQA & SAMSum & PCount & PR-en & Lcc & RB-P &  \\

\midrule
\multicolumn{18}{c}{Qwen3-30B-A3B-Instruct-2507, 20\% Cache} \\
\midrule
FullKV & 27.71 & 40.09 & 53.83 & 66.99 & 62.1 & 32.69 & 30.56 & 22.24 & 24.25 & 76 & 88.82 & 48.17 & 14 & 100 & 74.62 & 70.8 & 52.05 \\
% SnapKV  & 11.02 & 30.06 & 28.14 & 50.65 & 39.4 & 17.49 & 27.23 & 18.26 & 20.79 & 66 & 88.96 & 46.65 & 12 & 78.5 & 73.78 & 67.92 & 42.31 \\
AdaKV   & 27.54 & 31.19 & 30.39 & 48 & 39.67 & 18.38 & 25.81 & 18.58 & 19.2 & 45 & 75.58 & 31.11 & 11.09 & 53 & 36.81 & 41.78 & 34.57 \\
% CAKE    & 12.86 & 15.86 & 19.78 & 31.39 & 27.08 & 8.39 & 18.48 & 14.59 & 17.47 & 26 & 87.13 & 37.83 & 2 & 5 & 67.04 & 69.02 & 28.75 \\
CriticalKV    & 32.06 & 30.79 & 33.19 & 54.06 & \textbf{47.49} & \textbf{26.13} & \textbf{29.08} & 19.88 & 21.04 & 70 & \textbf{88.24} & 47.02 & \textbf{12} & 98.5 & 75.34 & 69.79 & 47.16 \\
LaProx    & \textbf{32.94} & \textbf{37.41} & \textbf{46.1} & \textbf{59.49} & 47.43 & 25.38 & 28.82 & \textbf{19.97} & \textbf{22.07} & \textbf{78} & 87.29 & \textbf{48.54} & \textbf{12} & \textbf{100} & \textbf{75.78} & \textbf{71.74} & \textbf{49.56} \\

\bottomrule
\end{tabular}
}
\end{table*}

\begin{table*}[ht]
\caption{ Comparison across 13 Ruler datasets. The best result is highlighted in \textbf{bold}.}
\centering
\small
\setlength{\tabcolsep}{3pt}
\label{tab:moe-ruler}

\resizebox{\textwidth}{!}{
\begin{tabular}{lcccccccccccccc}
\toprule
Method & CWE & FWE & NIAH\_MK1 & NIAH\_MK2 & NIAH\_MK3 & NIAH\_MQ & NIAH\_MV & NIAH\_S1 & NIAH\_S2 & NIAH\_S3 & QA1 & QA2 & VT & Avg. \\
\midrule
\multicolumn{15}{c}{Qwen3-30B-A3B-Instruct-2507, 20\% Cache} \\
\midrule
FullKV & 83.2 & 99.33 & 100 & 100 & 100 & 100 & 98.5 & 100 & 100 & 100 & 86 & 64 & 100 & 94.69
 \\
AdaKV  & 57.8 & 92 & 8 & 10 & 2 & 16 & 0 & 16 & 0 & 4 & 32 & 46 & 28.4 & 24.02 \\
CriticalKV   & 60.6 & 96 & \textbf{100} & 14 & 2 & 100 & \textbf{98} & 100 & 100 & 4 & 34 & 48 & 97.2 & 65.67 \\
LaProx   & \textbf{64.5} & \textbf{97} & \textbf{100} & \textbf{100} & \textbf{99} & \textbf{100} & 92.75 & \textbf{100} & \textbf{100} & \textbf{100} & \textbf{71} & \textbf{63} & \textbf{99.8} & \textbf{91.31} \\
\bottomrule
\end{tabular}
}
\end{table*}

The results in Tables \ref{tab:moe-longbench} and \ref{tab:moe-ruler} further demonstrate the effectiveness of our method in the MoE setting, where it limits performance degradation to only around 3%, substantially outperforming existing baselines.

\subsection{Scalability to Larger Models}
\label{appendix:large_model}

Section \ref{appendix:moe} already includes experiments on a larger model,
Qwen3-30B-A3B-Instruct-2507 MoE, where LaProx consistently outperforms the
strongest baselines. These results demonstrate that the proposed scoring
mechanism generalizes beyond the 7B--8B models evaluated in the main paper.

To further strengthen this evidence, we additionally evaluate LaProx on the
Qwen3-32B dense model, comparing against the same baselines, AdaKV and
CriticalKV, under the same setting. The results are shown below.

\begin{table}[thb]
    \centering
    \caption{LongBench results on Qwen3-32B with a 20\% cache size.}
    \label{tab:qwen32b_longbench}
    \resizebox{\linewidth}{!}{%
    \begin{tabular}{lccccccccccccccccc}
        \toprule
        % Method & NrtvQA & Qasper & MF & Hotpot & 2WikiQA & MuSiQue & GovRep & QMSum & MultiNews & TREC & TriviaQA & SAMSum & PCount & PR & LCC & RB-P & Avg. \\
        % \midrule
        % AdaKV & 24.39 & 32.96 & 28.36 & 55.00 & 44.90 & 23.09 & 28.67 & 20.32 & 22.49 & 52.50 & \textbf{89.82} & \textbf{42.39} & \textbf{15} & 86 & \textbf{10.23} & 30.81 & 37.93 \\

        % CriticalKV & 27.34 & 37.24 & 32.11 & 56.61 & 49.67 & \textbf{36.96} & 30.37 & 21.54 & 23.08 & 52.50 & \textbf{89.82} & 41.45 & \textbf{15} & \textbf{100} & 7.80 & \textbf{35.87} & 41.08 \\

        % LaProx & \textbf{27.37} & \textbf{42.65} & \textbf{49.48} & \textbf{62.06} & \textbf{53.81} & 35.22 & \textbf{31.01} & \textbf{23.47} & \textbf{25.13} & \textbf{71.00} & \textbf{89.82} & 41.33 & \textbf{15} & \textbf{100} & 8.46 & 33.06 & \textbf{44.29} \\
        Methods & Single-Doc QA & Multi-Doc QA & Summarization & Few-shot Learning & Synthetic & Code & Average \\
        \midrule
        AdaKV & 28.57 & 41.00 & 23.83 & 61.57 & 50.50 & 20.52 & 37.93 \\

        CriticalKV & 32.23 & 47.75 & 25.00 & 61.26 & \textbf{57.50} & \textbf{21.83} & 41.08 \\

        LaProx & \textbf{39.84} & \textbf{50.37} & \textbf{26.54} & \textbf{67.39} & \textbf{57.50} & 20.76 & \textbf{44.29} \\
        \bottomrule
    \end{tabular}%
    }
\end{table}

\begin{table}[thb]
    \centering
    \caption{RULER results on Qwen3-32B with a 20\% cache size.}
    \label{tab:qwen32b_ruler}
    \resizebox{\linewidth}{!}{%
    \begin{tabular}{lccccc}
        \toprule
        Methods & Word Extraction & Needle-in-A-Haystack & Multi-Hop Tracing & Question Answering &
        Average \\
        \midrule
        AdaKV & 79.57 & 20.38 & 79.20 & 36 & 36.41 \\

        CriticalKV & 71.80 & 58.25 & 99.60 & 45 & 61.48 \\

        LaProx & \textbf{86.47} & \textbf{84.57} & \textbf{100} & \textbf{65} & \textbf{84.23} \\
        \bottomrule
    \end{tabular}}
\end{table}

As shown in Tables~\ref{tab:qwen32b_longbench}
and~\ref{tab:qwen32b_ruler}, \textbf{LaProx consistently maintains its advantage on
both dense and MoE models at the 30B+ scale.} On LongBench, it achieves the
highest scores on the majority of tasks. The improvement is even more
pronounced on RULER, particularly on challenging tasks such as
Needle-in-A-Haystack, where LaProx substantially outperforms both AdaKV and
CriticalKV.

These experiments demonstrate that the proposed method scales well beyond the
7B--8B models presented in the main paper and remains effective for larger
models.

\subsection{Detailed Performance Across Cache Budgets on LongBench Benchmark}
Tables \ref{table-llama}, \ref{table-mistral}, and \ref{table-qwen} report detailed LongBench results shown in Figure \ref{fig:longbench} for LaProx and competing baselines under cache budgets ranging from 128 to 1024 tokens on Meta-Llama-3.1-8B-Instruct, Mistral-7B-Instruct-v0.3, and Qwen3-8B, respectively.

Overall, the results show that LaProx consistently outperforms prior methods across all LongBench task categories on all evaluated models.

\begin{table*}[tb]
\caption{Comparison across 16 LongBench datasets on Meta-Llama-3.1-8B-Instruct for cache budgets from 128L to 1024L. The best result is highlighted in \textbf{bold} and the second best in \underline{underline}.}
\label{table-llama}
\centering
\small
\setlength{\tabcolsep}{3pt}
\resizebox{\textwidth}{!}{
\begin{tabular}{l ccc ccc ccc ccc cc cc c}
\toprule
\multirow{2}{*}{Method} & \multicolumn{3}{c}{Single-Document QA} & \multicolumn{3}{c}{Multi-Document QA} & \multicolumn{3}{c}{Summarization} & \multicolumn{3}{c}{Few-shot Learning} & \multicolumn{2}{c}{Synthetic} & \multicolumn{2}{c}{Code} & \multirow{2}{*}{Avg.} \\
\cmidrule(lr){2-4} \cmidrule(lr){5-7} \cmidrule(lr){8-10} \cmidrule(lr){11-13} \cmidrule(lr){14-15} \cmidrule(lr){16-17}

 & NrtvQA & Qasper & MF-en & HotpotQA & 2WikiMQA & Musique & GovReport & QMSum & MultiNews & TREC & TriviaQA & SAMSum & PCount & PR-en & Lcc & RB-P &  \\

\midrule
\multicolumn{18}{c}{Meta-Llama-3.1-8B-Instruct, $B_{\text{total}} = Full$} \\
\midrule
FullKV & 30.28 & 45.46 & 54.94 & 56.02 & 46.5 & 31.28 & 35.11 & 25.28 & 27.19 & 72.5 & 91.65 & 43.86 & 8.88 & 99.5 & 65.13 & 58.7 & 49.51 \\
\midrule
\multicolumn{18}{c}{Meta-Llama-3.1-8B-Instruct, $B_{\text{total}} = 128L$} \\
\midrule
SLLM & 22.45 & 20.83 & 31.74 & 44.86 & 39.31 & 23.88 & 18.75 & 20.78 & 18.57 & 40.5 & 85.57 & 38.24 & 8.16 & \textbf{99.5} & 58.89 & 49.7 & 38.86 \\
SnapKV & 25.53 & 23.92 & 48.15 & \underline{52.94} & 43.5 & 26.67 & 19.48 & 22.47 & 19.68 & 47 & 89.56 & 39.51 & \underline{8.41} & 98 & 59.43 & 52.84 & 42.31 \\
AdaKV & 25.74 & 24.8 & 49.38 & 52.73 & 43.75 & 27.8 & 20.37 & \underline{23.02} & \underline{20.38} & \underline{49} & \textbf{90.19} & 40.12 & 8.38 & \underline{99} & \underline{60.92} & \underline{54.4} & \underline{43.12} \\
CAKE & \textbf{25.93} & 25.5 & 48.33 & 51.95 & 43.42 & \underline{28.94} & \underline{20.91} & 22.93 & 19.96 & 47.5 & 88.84 & \textbf{41.26} & 8.38 & \underline{99} & 59.56 & 51.97 & 42.77 \\
CriticalKV & 24.56 & \underline{25.89} & \underline{49.86} & 52.42 & \underline{43.89} & 27.96 & 20.71 & 22.69 & 20.06 & 48.5 & 89.61 & 39.94 & 8.38 & \underline{99} & 60.08 & 52.56 & 42.88 \\
LaProx & \underline{25.78} & \textbf{31.38} & \textbf{52.97} & \textbf{55.17} & \textbf{44.95} & \textbf{29.46} & \textbf{21.66} & \textbf{23.66} & \textbf{21.47} & \textbf{59} & \underline{89.94} & \underline{41.09} & \textbf{8.89} & \textbf{99.5} & \textbf{61.91} & \textbf{56.16} & \textbf{45.19} \\

\midrule
\multicolumn{18}{c}{Meta-Llama-3.1-8B-Instruct, $B_{\text{total}} = 256L$} \\
\midrule
SLLM     & 22.52 & 23.45 & 32.21 & 45.69 & 37.93 & 24.61 & 21.12 & 20.86 & 20.91 & 46 & 87.5 & 41 & 7.61 & \textbf{99.5} & 61.11 & 51.44 & 40.22 \\
SnapKV  & 27.6 & 33.56 & 51.53 & 54.95 & 44.8 & 29.27 & 22.56 & 23.44 & 22.28 & 55 & 91.38 & 40.69 & \textbf{8.88} & \textbf{99.5} & 62.48 & 55.5 & 45.21 \\
AdaKV   & 26.55 & 34.38 & \underline{52.96} & \underline{55.21} & 44.23 & 29.24 & 23.12 & \underline{24} & \underline{22.85} & \underline{60.5} & \underline{91.57} & 41.04 & 7.84 & \textbf{99.5} & \underline{63.32} & \underline{56.72} & \underline{45.81} \\
CAKE    & 27.28 & \underline{34.41} & 52.34 & 54.14 & \underline{45.57} & \underline{30.06} & \underline{23.66} & \underline{24} & 22.63 & 59 & 91.25 & 41.48 & 7.97 & \textbf{99.5} & 62.1 & 54.37 & 45.61 \\
CriticalKV   & \textbf{28.31} & 33.76 & 51.49 & 54.96 & 44.09 & 29.73 & 23.39 & \textbf{24.14} & 22.6 & 57.5 & 91.49 & \underline{41.75} & 8.44 & \textbf{99.5} & 62.72 & 56.49 & 45.64 \\
LaProx    & \underline{28.02} & \textbf{38.56} & \textbf{53.32} & \textbf{55.23} & \textbf{45.72} & \textbf{30.83} & \textbf{24.84} & 23.78 & \textbf{23.92} & \textbf{67.5} & \textbf{92.12} & \textbf{42.4} & \underline{8.56} & \textbf{99.5} & \textbf{63.57} & \textbf{57.61} & \textbf{47.22} \\

\midrule
\multicolumn{18}{c}{Meta-Llama-3.1-8B-Instruct, $B_{\text{total}} = 512L$} \\
\midrule
SLLM     & 25.65 & 26.05 & 33.92 & 46.14 & 36.98 & 24.69 & 23.87 & 21.2 & 23.62 & 57.5 & 87.7 & 41.92 & 7.38 & \underline{96.5} & 62.37 & 53.72 & 41.83 \\
SnapKV  & 29.7 & 39.31 & 53.78 & 55.09 & \underline{45.95} & 30.44 & 25.64 & 24.16 & 24.14 & 65.5 & 91.66 & 41.18 & 8.12 & \textbf{99.5} & \underline{64.7} & 56.9 & 47.23 \\
AdaKV   & 28.54 & 41.77 & 54.18 & \textbf{55.68} & 45.44 & 30.48 & 25.77 & 24.34 & \underline{25} & \underline{69} & \textbf{92.04} & 41.66 & 7.66 & \textbf{99.5} & \textbf{64.75} & \underline{57.66} & \underline{47.71} \\
CAKE    & 29.59 & \underline{41.8} & \textbf{54.39} & 54.94 & 45.79 & \underline{30.91} & \underline{26.16} & \underline{24.44} & 24.83 & 63.5 & 91.93 & 41.8 & \underline{8.17} & \textbf{99.5} & 63.99 & 56.82 & 47.41 \\
CriticalKV   & \textbf{30.33} & 39.69 & 53.81 & 54.91 & 45.89 & \textbf{31.05} & 25.59 & \textbf{24.74} & 24.44 & 62 & \underline{91.94} & \textbf{42.66} & 8.12 & \textbf{99.5} & 64.53 & 57.24 & 47.27 \\
LaProx    & \underline{29.82} & \textbf{43.46} & \underline{54.33} & \underline{55.44} & \textbf{46.16} & 30.83 & \textbf{27.08} & 24.1 & \textbf{25.3} & \textbf{70.5} & 91.89 & \underline{42.48} & \textbf{8.4} & \textbf{99.5} & 64.1 & \textbf{58.16} & \textbf{48.23} \\

\midrule
\multicolumn{18}{c}{Meta-Llama-3.1-8B-Instruct, $B_{\text{total}} = 1024L$} \\
\midrule
SLLM     & 25.34 & 29.79 & 37.08 & 47.79 & 39.83 & 24.44 & 26.28 & 21.43 & 25.85 & 63.5 & 88.83 & 42.56 & 7.91 & \underline{88.5} & 63.17 & 55.96 & 43.02 \\
SnapKV  & 30.54 & \underline{43.93} & 53.96 & \underline{55.8} & 46.3 & \underline{31.19} & 28.41 & \textbf{24.75} & 25.99 & 69.5 & 91.78 & 42.22 & \underline{8.16} & \textbf{99.5} & 64.62 & \underline{58.85} & 48.46 \\
AdaKV   & \textbf{31.09} & 43.71 & \underline{54.44} & \textbf{55.82} & 46.15 & 30.65 & 28.54 & 24.41 & 25.82 & \underline{71} & 91.87 & 42.32 & 7.96 & \textbf{99.5} & 64.57 & 58.56 & \underline{48.52} \\
CAKE    & 30.32 & 43.25 & 53.98 & 55.69 & \underline{46.53} & 30.96 & \underline{28.81} & 24.21 & 25.74 & 68 & \textbf{91.9} & \underline{42.87} & 7.79 & \textbf{99.5} & \textbf{65.01} & 58.11 & 48.29 \\
CriticalKV   & 30.14 & 43.8 & 54.11 & 55.63 & \textbf{46.6} & 30.88 & 28.76 & \underline{24.66} & \underline{26.1} & 68.5 & \underline{91.89} & 42.42 & 7.93 & \textbf{99.5} & \underline{64.83} & \textbf{58.9} & 48.41 \\
LaProx    & \underline{31.08} & \textbf{44.45} & \textbf{55.64} & \textbf{55.82} & 45.98 & \textbf{31.96} & \textbf{29.22} & 24.42 & \textbf{26.36} & \textbf{71.5} & 91.75 & \textbf{43.32} & \textbf{8.23} & \textbf{99.5} & \textbf{65.01} & 58.69 & \textbf{48.94} \\

\bottomrule
\end{tabular}
}
\vskip -0.1in
\end{table*}

\begin{table*}[tb]
\caption{Comparison across 16 LongBench datasets on Mistral-7B-Instruct-v0.3 for cache budgets from 128L to 1024L. The best result is highlighted in \textbf{bold} and the second best in \underline{underline}.}
\label{table-mistral}
\centering
\small
\setlength{\tabcolsep}{3pt}
\resizebox{\textwidth}{!}{
\begin{tabular}{l ccc ccc ccc ccc cc cc c}
\toprule
\multirow{2}{*}{Method} & \multicolumn{3}{c}{Single-Document QA} & \multicolumn{3}{c}{Multi-Document QA} & \multicolumn{3}{c}{Summarization} & \multicolumn{3}{c}{Few-shot Learning} & \multicolumn{2}{c}{Synthetic} & \multicolumn{2}{c}{Code} & \multirow{2}{*}{Avg.} \\
\cmidrule(lr){2-4} \cmidrule(lr){5-7} \cmidrule(lr){8-10} \cmidrule(lr){11-13} \cmidrule(lr){14-15} \cmidrule(lr){16-17}

 & NrtvQA & Qasper & MF-en & HotpotQA & 2WikiMQA & Musique & GovReport & QMSum & MultiNews & TREC & TriviaQA & SAMSum & PCount & PR-en & Lcc & RB-P &  \\

\midrule
\multicolumn{18}{c}{Mistral-7B-Instruct-v0.3, $B_{\text{total}} = Full$} \\
\midrule
FullKV & 29.07 & 41.58 & 52.88 & 49.37 & 39.01 & 28.58 & 34.81 & 25.66 & 27.82 & 76 & 88.59 & 47.4 & 5.5 & 98 & 61.4 & 62.53 & 48.01 \\
\midrule
\multicolumn{18}{c}{Mistral-7B-Instruct-v0.3, $B_{\text{total}} = 128L$} \\
\midrule
SLLM & 21.42 & 22.28 & 26.82 & 37.28 & 33.31 & 17.61 & 16.73 & 19.77 & 17.86 & 45.5 & 85.64 & 40.47 & 5.5 & 80 & 55.13 & 52.07 & 36.08 \\
SnapKV       & 23.81 & 25.05 & 46.92 & 44.98 & 35.06 & 23.11 & 20.49 & 21.55 & 19.32 & 43.5 & \textbf{89.78} & 42.87 & 6 & 93.5 & 55.95 & 54.6 & 40.41 \\
AdaKV        & 24.08 & 25.52 & 47.08 & 47.16 & 35.16 & 23.83 & 20.23 & 22.28 & 20.7 & 45 & 88.92 & \textbf{43.77} & \underline{6.5} & 92.5 & \underline{56.63} & \underline{55.81} & 40.95 \\
CAKE         & 25.04 & \underline{27.32} & \underline{48.49} & \underline{47.48} & 35.24 & \underline{24.63} & \textbf{21.32} & \underline{22.54} & \underline{20.78} & 45.5 & 89.41 & 43.37 & \textbf{7} & \underline{95} & 56.47 & 54.99 & \underline{41.54} \\
CriticalKV   & \underline{25.42} & 25.95 & 46.43 & 46.44 & \underline{35.92} & 23.93 & 21.12 & 22.35 & 19.26 & \underline{47} & 89.23 & 43.62 & 6.5 & 93.5 & 55.99 & 54.51 & 41.07 \\
LaProx         & \textbf{28} & \textbf{30.26} & \textbf{54.02} & \textbf{47.86} & \textbf{37.54} & \textbf{25.54} & \underline{21.26} & \textbf{23.31} & \textbf{22.1} & \textbf{63.5} & \underline{89.7} & \underline{43.46} & 5 & \textbf{96} & \textbf{58.85} & \textbf{57.66} & \textbf{44.00} \\

\midrule
\multicolumn{18}{c}{Mistral-7B-Instruct-v0.3, $B_{\text{total}} = 256L$} \\
\midrule
SLLM      & 22.49 & 23.21 & 29.75 & 39.62 & 32.51 & 16.96 & 19.14 & 19.29 & 20.12 & 54.5 & 85.12 & 42.98 & \underline{5.5} & 80 & 57.71 & 55.05 & 37.75 \\
SnapKV   & \underline{27.33} & 31.02 & 51.54 & \underline{48.95} & 36.54 & 27.04 & 22.16 & 22.94 & 22.06 & 55.5 & \underline{89.4} & \underline{44.52} & 5 & \textbf{96.5} & 58.87 & 58.03 & 43.58 \\
AdaKV    & 27.01 & 31.11 & 52.32 & 47.14 & 37.06 & \underline{28.01} & 22.45 & 23.05 & 22.65 & \textbf{62.5} & \textbf{89.44} & 44.31 & \textbf{6} & \textbf{96.5} & 58.65 & 59.32 & \underline{44.22} \\
CAKE     & 27.1 & \underline{31.42} & \underline{53.3} & \textbf{49.38} & \underline{37.56} & 27.63 & \underline{22.87} & \underline{23.17} & \underline{23.08} & 57 & 89.24 & \textbf{44.57} & 4.5 & \textbf{96.5} & \textbf{59.5} & \underline{59.47} & 44.14 \\
CriticalKV   & 26.8 & 31.32 & 52.35 & 48.85 & 36.85 & 27.33 & 22.55 & 23.1 & 22.95 & \underline{58.5} & 89.23 & 43.96 & 4.5 & \textbf{96.5} & \underline{59.18} & 59.38 & 43.96 \\
LaProx     & \textbf{28.53} & \textbf{34.71} & \textbf{54.87} & 48.88 & \textbf{37.62} & \textbf{28.99} & \textbf{23.53} & \textbf{24.07} & \textbf{23.64} & \textbf{62.5} & 89.03 & 44.29 & \underline{5.5} & \underline{96} & 59.15 & \textbf{59.95} & \textbf{45.08} \\

\midrule
\multicolumn{18}{c}{Mistral-7B-Instruct-v0.3, $B_{\text{total}} = 512L$} \\
\midrule
SLLM      & 24.19 & 25.89 & 30.45 & 40.6 & 32.36 & 17.35 & 22.02 & 20.2 & 23.28 & 65.5 & 86.95 & 43.75 & \textbf{6} & 81 & 59.29 & 56.34 & 39.70 \\
SnapKV   & 28.08 & 33.93 & \textbf{53.91} & 48.96 & 37.63 & 27.39 & 24 & 23.77 & 24.07 & 67 & 89.23 & 45.26 & 5 & 96.5 & 59.95 & 60.22 & 45.31 \\
AdaKV    & \underline{29.04} & 34.64 & 53.35 & 48.99 & 36.72 & 27.66 & 24.41 & 23.99 & 24.47 & \underline{71} & 89.03 & \underline{45.69} & 5 & \underline{97} & 60.03 & 60.96 & 45.74 \\
CAKE     & 28.68 & 35.54 & 53.68 & 48.64 & \underline{38.6} & \underline{28.32} & \underline{25.07} & 23.76 & \textbf{25.05} & 69.5 & \underline{89.44} & 45.64 & \underline{5.5} & \textbf{97.5} & 59.98 & 60.87 & \underline{45.98} \\
CriticalKV   & 28.09 & \underline{37.07} & \underline{53.71} & \underline{49.52} & 37.34 & 28.08 & 24.48 & \underline{24.25} & 24.84 & \underline{71} & 89.33 & 44.28 & 3.5 & \textbf{97.5} & \underline{60.33} & \underline{61.33} & 45.91 \\
LaProx     & \textbf{29.7} & \textbf{37.29} & 52.9 & \textbf{49.57} & \textbf{38.73} & \textbf{28.47} & \textbf{25.41} & \textbf{25.09} & \underline{24.77} & \textbf{74.5} & \textbf{89.66} & \textbf{45.94} & \textbf{6} & \underline{97} & \textbf{60.83} & \textbf{61.86} & \textbf{46.74} \\

\midrule
\multicolumn{18}{c}{Mistral-7B-Instruct-v0.3, $B_{\text{total}} = 1024L$} \\
\midrule
SLLM      & 24.79 & 27.88 & 30.99 & 42.91 & 32.65 & 18.03 & 24.64 & 20.7 & 25.45 & 68.5 & 88.71 & 45.37 & \underline{5.5} & 82.5 & 61.1 & 59.24 & 41.19 \\
SnapKV   & 28.35 & 37.47 & \underline{53.41} & 49.12 & \textbf{39.46} & \textbf{28.38} & 26.29 & 24.67 & 25.94 & 71 & 88.89 & \underline{46.84} & 5 & 97.5 & 61.08 & 61.77 & 46.57 \\
AdaKV    & 28.14 & 37.35 & 52.86 & 48.97 & 38.59 & 28.24 & 26.6 & \underline{24.82} & 25.81 & 72.5 & \underline{89.19} & 46.12 & 5 & \underline{98.5} & 60.82 & \underline{62.36} & 46.61 \\
CAKE     & \textbf{29.93} & \underline{37.6} & 53.12 & \underline{50.3} & 38.63 & \underline{28.36} & \underline{27.43} & 24.49 & \textbf{26.73} & 73 & \underline{89.19} & 45.95 & \textbf{6} & 97.5 & 61.01 & 62.13 & \underline{46.96} \\
CriticalKV   & \underline{29.06} & 37.28 & \textbf{53.9} & 49.92 & 38.24 & 28.21 & 26.62 & 24.14 & \underline{26.56} & \underline{73.5} & \underline{89.19} & 45.27 & 5 & 97.5 & \textbf{61.69} & \textbf{62.67} & 46.79 \\
LaProx     & 28.99 & \textbf{39.16} & 52.1 & \textbf{50.39} & \underline{38.87} & 28.05 & \textbf{27.77} & \textbf{25.12} & 26.36 & \textbf{76} & \textbf{89.61} & \textbf{47} & \underline{5.5} & \textbf{99} & \underline{61.47} & 62.31 & \textbf{47.36} \\

\bottomrule
\end{tabular}
}
\vskip -0.1in
\end{table*}

\begin{table*}[tb]
\caption{Comparison across 16 LongBench datasets on Qwen3-8B for cache budgets from 128L to 1024L. The best result is highlighted in \textbf{bold} and the second best in \underline{underline}.}
\label{table-qwen}
\centering
\small
\setlength{\tabcolsep}{3pt}
\resizebox{\textwidth}{!}{
\begin{tabular}{l ccc ccc ccc ccc cc cc c}
\toprule
\multirow{2}{*}{Method} & \multicolumn{3}{c}{Single-Document QA} & \multicolumn{3}{c}{Multi-Document QA} & \multicolumn{3}{c}{Summarization} & \multicolumn{3}{c}{Few-shot Learning} & \multicolumn{2}{c}{Synthetic} & \multicolumn{2}{c}{Code} & \multirow{2}{*}{Avg.} \\
\cmidrule(lr){2-4} \cmidrule(lr){5-7} \cmidrule(lr){8-10} \cmidrule(lr){11-13} \cmidrule(lr){14-15} \cmidrule(lr){16-17}

 & NrtvQA & Qasper & MF-en & HotpotQA & 2WikiMQA & Musique & GovReport & QMSum & MultiNews & TREC & TriviaQA & SAMSum & PCount & PR-en & Lcc & RB-P &  \\

\midrule
\multicolumn{18}{c}{Qwen3-8B, $B_{\text{total}} = Full$} \\
\midrule
FullKV       & 32.26 & 46.59 & 52.73 & 59.35 & 51.21 & 33.2 & 32.44 & 24.13 & 25.68 & 71 & 65.82 & 40 & 1 & 100 & 53.06 & 50.17 & 46.17 \\
\midrule
\multicolumn{18}{c}{Qwen3-8B, $B_{\text{total}} = 128L$} \\
\midrule
SLLM          & 12.18 & 28.05 & 21.79 & 36.34 & 39.07 & 6.88 & 15.78 & 18.54 & 15.95 & 43 & 62.56 & 33.3 & \underline{3} & 73 & 45.16 & 45.43 & 31.25 \\
SnapKV       & 16.34 & 31.35 & 42.37 & 52.67 & 42.33 & 18.26 & 16.03 & 19.68 & 16.49 & 53 & \underline{75.69} & 34.44 & 1 & \textbf{99} & 46.86 & 47.86 & 38.33 \\
AdaKV        & \textbf{21.66} & 32.58 & 43.96 & 50.81 & 42.95 & \underline{20.9} & 16.45 & 19.33 & 16.02 & 51 & 73.92 & 34.24 & 1 & \underline{98} & 45.77 & 46.92 & 38.47 \\
CAKE         & \underline{19.69} & 32.67 & 45.07 & \underline{57.5} & \textbf{48.02} & 20.5 & \underline{17.82} & \underline{20.87} & 16.73 & 47 & 69.51 & \underline{35.39} & \textbf{5} & \textbf{99} & 45.55 & 45.94 & 39.14 \\
CriticalKV   & 18.47 & \underline{35.83} & \underline{45.78} & 48.59 & 42.84 & 19.09 & 16.67 & 19.94 & \textbf{17.61} & \underline{56} & 75.31 & 35.3 & 0 & \textbf{99} & \underline{47.79} & \underline{48.28} & \underline{39.15} \\
LaProx         & 17.51 & \textbf{36.91} & \textbf{51.22} & \textbf{58.64} & \underline{46.31} & \textbf{23.85} & \textbf{18.68} & \textbf{21.21} & \underline{17.32} & \textbf{61} & \textbf{76.49} & \textbf{35.86} & 1 & \textbf{99} & \textbf{48.86} & \textbf{51.21} & \textbf{41.57} \\

\midrule
\multicolumn{18}{c}{Qwen3-8B, $B_{\text{total}} = 256L$} \\
\midrule
SLLM     & 14.01 & 28.83 & 22.05 & 36.99 & 38.97 & 7.83 & 18.6 & 18.7 & 18.73 & 52 & 70.4 & 34.79 & \underline{5} & 67 & 49.86 & 46.88 & 33.16 \\
SnapKV  & 18.2 & 37.54 & 49.44 & \underline{63.13} & \textbf{48.93} & 24.45 & 20.4 & 21.7 & 19.2 & 59 & 70.62 & 37.04 & 0 & \textbf{100} & 50.64 & \textbf{53.17} & 42.09 \\
AdaKV   & \underline{22.31} & 36.05 & 47.75 & 57.07 & 44.62 & 22.33 & \underline{20.47} & 20.46 & 18.83 & 64 & \textbf{74.07} & 37.82 & 1 & \textbf{100} & 50.12 & 49.91 & 41.67 \\
CAKE    & 20.37 & 38.78 & \underline{50.31} & \textbf{64.01} & \underline{48.27} & 25.47 & 22.02 & 21.56 & 19.81 & 61 & \underline{73.17} & \underline{38.12} & \textbf{9} & \textbf{100} & 49.14 & 49.66 & \underline{43.16} \\
CriticalKV   & 22.15 & \underline{40.41} & 48.89 & 58.01 & 46.83 & \underline{27.89} & 21.62 & \underline{21.72} & \underline{20.1} & \underline{68} & 70.52 & 37.64 & 0 & \textbf{100} & \underline{51.35} & 52.14 & 42.97 \\
LaProx    & \textbf{23.5} & \textbf{41.26} & \textbf{50.96} & 62.98 & 46.55 & \textbf{30.44} & \textbf{22.55} & \textbf{22.29} & \textbf{21.63} & \textbf{70} & 72.97 & \textbf{38.17} & 0 & \textbf{100} & \textbf{52.51} & \underline{53} & \textbf{44.30} \\

\midrule
\multicolumn{18}{c}{Qwen3-8B, $B_{\text{total}} = 512L$} \\
\midrule
SLLM     & 15.13 & 31.56 & 24.92 & 39.07 & 41.29 & 8.16 & 22.21 & 18.87 & 22.25 & 62 & 74.21 & 35.3 & \textbf{7} & \underline{51} & 52.22 & 50.92 & 34.76 \\
SnapKV  & 22.11 & 42.59 & \textbf{53.05} & \underline{61.28} & 49.68 & \underline{30.65} & 23.94 & \underline{23.26} & 21.88 & 67 & 69.27 & \textbf{39.44} & 0 & \textbf{100} & \underline{54.12} & 53.16 & 44.46 \\
AdaKV   & 22.56 & 40.55 & 48.46 & 60.76 & 47.57 & 30.57 & 23.93 & 21.88 & 21.47 & 67 & 68.97 & 38.13 & 0 & \textbf{100} & 52.28 & 52.52 & 43.54 \\
CAKE    & 23.06 & 43.77 & 51.53 & 60.53 & \underline{49.84} & \textbf{31.84} & 25.25 & 23.11 & \underline{22.76} & 64 & \textbf{75.43} & 38.42 & \underline{6} & \textbf{100} & 53.02 & 53.45 & \underline{45.12} \\
CriticalKV   & \textbf{25.47} & \underline{43.98} & 51.37 & 60.7 & 48.81 & 28.44 & \textbf{25.5} & 22.75 & 22.67 & \underline{69} & 71.47 & \underline{38.47} & 0 & \textbf{100} & \textbf{54.89} & \underline{53.75} & 44.82 \\
LaProx    & \underline{24.76} & \textbf{45.84} & \underline{52.68} & \textbf{61.96} & \textbf{49.98} & 28.87 & \underline{25.37} & \textbf{23.3} & \textbf{24.33} & \textbf{71} & \underline{74.77} & 38.3 & 0 & \textbf{100} & 53.8 & \textbf{53.76} & \textbf{45.55} \\

\midrule
\multicolumn{18}{c}{Qwen3-8B, $B_{\text{total}} = 1024L$} \\
\midrule
SLLM     & 19.18 & 33.34 & 27.03 & 44.7 & 40.99 & 9.59 & 25.6 & 20.15 & 21.02 & 63 & 80.02 & 37.23 & \textbf{7} & \underline{39} & 53.29 & 50.53 & 35.73 \\
SnapKV  & \underline{26.38} & 43.49 & \textbf{53.2} & \textbf{62.25} & 48.92 & 29.77 & 27.31 & \textbf{23.49} & 20.87 & \underline{69} & 71.97 & 38.09 & \underline{1} & \textbf{100} & 54.51 & 51.95 & 45.13 \\
AdaKV   & 24.88 & 42.59 & 51.68 & 60.18 & 46.78 & 30.56 & 26.68 & 21.96 & 20.88 & \textbf{70} & \underline{74.77} & \underline{38.43} & 0 & \textbf{100} & 53.28 & 51.42 & 44.63 \\
CAKE    & 24.09 & \underline{44.97} & 52.02 & 60.58 & \textbf{50.62} & \textbf{31.54} & \underline{27.83} & \underline{23.25} & \textbf{24.33} & \textbf{70} & 73.99 & 37.71 & \underline{1} & \textbf{100} & \underline{54.67} & \underline{53.56} & \underline{45.63} \\
CriticalKV   & 24.97 & 43.61 & 51.95 & 60.02 & 48.73 & 28.8 & \textbf{28.23} & 23 & \underline{21.28} & \textbf{70} & 73.7 & 37.59 & 0 & \textbf{100} & \textbf{54.71} & \textbf{53.8} & 45.02 \\
LaProx    & \textbf{27.71} & \textbf{46.12} & \underline{53.04} & \underline{61.83} & \underline{50.48} & \underline{30.72} & 27.58 & 22.9 & 21.07 & \textbf{70} & \textbf{76.27} & \textbf{41.35} & \underline{1} & \textbf{100} & 54.18 & 52.37 & \textbf{46.04} \\

\bottomrule
\end{tabular}
}
\vskip -0.1in
\end{table*}

\section{Design Choice Discussion}
The goal of KV-cache eviction is to identify which cached key-value (KV) pairs can be removed while minimizing their impact on model outputs. Since cached KV pairs are directly consumed only by the attention mechanism, their eviction primarily affects the attention output $AVW_O$, which is subsequently incorporated into the residual stream. Therefore, we formulate our optimization objective around the perturbation of $AVW_O$, as it \textbf{directly captures the impact of KV eviction} before its effects propagate through subsequent computations.

Although this perturbation propagates through residual connections, MLPs, and subsequent Transformer blocks, these components operate on the modified residual stream rather than individual KV pairs. Once the attention output is computed, the contributions of individual KV pairs have already been aggregated and can no longer be explicitly isolated. Consequently, directly estimating the effect of each KV pair on the final model output would require tracing its influence through all subsequent transformations. Such an approach would introduce substantial computational complexity, making it impractical for \textbf{efficient online KV-cache eviction.}

Furthermore, assigning importance scores to individual KV pairs based on their contributions to the attention computation is a fundamental \textbf{characteristic of efficient KV-cache eviction frameworks.} Within this formulation, LaProx provides a more principled theoretical framework for estimating the impact of KV eviction, which consistently translates into superior empirical performance.

%%%%%%%%%%%%%%%%%%%%%%%%%%%%%%%%%%%%%%%%%%%%%%%%%%%%%%%%%%%%

\clearpage
\section*{NeurIPS Paper Checklist}
\begin{enumerate}
\item {\bf Claims}
    \item[] Question: Do the main claims made in the abstract and introduction accurately reflect the paper's contributions and scope?
    \item[] Answer: \answerYes{}
    \item[] Justification: The statements presented in the abstract and introduction are completely consistent with the paper’s primary contributions.

\item {\bf Limitations}
    \item[] Question: Does the paper discuss the limitations of the work performed by the authors?
    \item[] Answer: \answerYes{}
    \item[] Justification: The limitations of our work are discussed in Appendix \ref{limitation}.

\item {\bf Theory assumptions and proofs}
    \item[] Question: For each theoretical result, does the paper provide the full set of assumptions and a complete (and correct) proof?
    \item[] Answer: \answerYes{}
    \item[] Justification: All theoretical results and statements are accompanied by complete proofs. Detailed proofs are provided in Appendix \ref{proof:MHA-decomposition} and \ref{appendix:layer-importance}.

    \item {\bf Experimental result reproducibility}
    \item[] Question: Does the paper fully disclose all the information needed to reproduce the main experimental results of the paper to the extent that it affects the main claims and/or conclusions of the paper (regardless of whether the code and data are provided or not)?
    \item[] Answer: \answerYes{}
    \item[] Justification: All experiment settings and implementation details are described in section \ref{sec:experiments} and \ref{appendix:imp}.

\item {\bf Open access to data and code}
    \item[] Question: Does the paper provide open access to the data and code, with sufficient instructions to faithfully reproduce the main experimental results, as described in supplemental material?
    \item[] Answer: \answerNo{}
    \item[] Justification: We provided pseudo codes for our implementations (Algorithm \ref{alg:kv_eviction} and \ref{alg:budget_allocation}) and the detailed experimental settings (Section \ref{sec:experiments} and Appendix \ref{appendix:imp}). We will open-source our code upon acceptance.

\item {\bf Experimental setting/details}
    \item[] Question: Does the paper specify all the training and test details (e.g., data splits, hyperparameters, how they were chosen, type of optimizer) necessary to understand the results?
    \item[] Answer: \answerYes{}
    \item[] Justification: Our method is a plug-and-play method that doesn't include any training process. Detailed settings are described in Section \ref{sec:experiments} and Appendix \ref{appendix:imp}.

\item {\bf Experiment statistical significance}
    \item[] Question: Does the paper report error bars suitably and correctly defined or other appropriate information about the statistical significance of the experiments?
    \item[] Answer: \answerNA{}
    \item[] Justification: We employ deterministic approaches without randomness, making error bars unnecessary. We also validate our approach through an exceptionally broad range of tests on LongBench, RULER, and InfiniteBench, which exceeds typical benchmarks and confirms the effectiveness of our method."

\item {\bf Experiments compute resources}
    \item[] Question: For each experiment, does the paper provide sufficient information on the computer resources (type of compute workers, memory, time of execution) needed to reproduce the experiments?
    \item[] Answer: \answerYes{}
    \item[] Justification: We detail the computation resources and efficiency in Section \ref{sec:efficiency}.
    
\item {\bf Code of ethics}
    \item[] Question: Does the research conducted in the paper conform, in every respect, with the NeurIPS Code of Ethics \url{https://neurips.cc/public/EthicsGuidelines}?
    \item[] Answer: \answerYes{}
    \item[] Justification: This work complies with the NeurIPS Code of Ethics.

\item {\bf Broader impacts}
    \item[] Question: Does the paper discuss both potential positive societal impacts and negative societal impacts of the work performed?
    \item[] Answer: \answerNA{}
    \item[] Justification: This paper improves LLMs inference performance and efficiency via a new KV cache management. The proposed method introduces no new ethical concerns beyond those already associated with LLMs. We do not identify any broader societal impacts requiring specific discussion.
    
\item {\bf Safeguards}
    \item[] Question: Does the paper describe safeguards that have been put in place for responsible release of data or models that have a high risk for misuse (e.g., pre-trained language models, image generators, or scraped datasets)?
    \item[] Answer: \answerNA{}
    \item[] Justification:  The paper does not release new models or data.

\item {\bf Licenses for existing assets}
    \item[] Question: Are the creators or original owners of assets (e.g., code, data, models), used in the paper, properly credited and are the license and terms of use explicitly mentioned and properly respected?
    \item[] Answer: \answerYes{}
    \item[] Justification: All third-party assets used in our work are properly credited in the paper with appropriate citations.

\item {\bf New assets}
    \item[] Question: Are new assets introduced in the paper well documented and is the documentation provided alongside the assets?
    \item[] Answer: \answerNA{}
    \item[] Justification: The paper does not release new assets.

\item {\bf Crowdsourcing and research with human subjects}
    \item[] Question: For crowdsourcing experiments and research with human subjects, does the paper include the full text of instructions given to participants and screenshots, if applicable, as well as details about compensation (if any)? 
    \item[] Answer: \answerNA{}
    \item[] Justification: The paper does not involve crowdsourcing nor research with human subjects.

\item {\bf Institutional review board (IRB) approvals or equivalent for research with human subjects}
    \item[] Question: Does the paper describe potential risks incurred by study participants, whether such risks were disclosed to the subjects, and whether Institutional Review Board (IRB) approvals (or an equivalent approval/review based on the requirements of your country or institution) were obtained?
    \item[] Answer: \answerNA{}
    \item[] Justification: The paper does not involve crowdsourcing nor research with human subjects.

\item {\bf Declaration of LLM usage}
    \item[] Question: Does the paper describe the usage of LLMs if it is an important, original, or non-standard component of the core methods in this research? Note that if the LLM is used only for writing, editing, or formatting purposes and does \emph{not} impact the core methodology, scientific rigor, or originality of the research, declaration is not required.
    %this research? 
    \item[] Answer: \answerNA{}
    \item[] Justification: We do not involve LLMs as any important, original, or non-standard components.

\end{enumerate}

\end{document}